\pdfoutput=1

\documentclass[11pt]{article}

\usepackage[]{acl}

\usepackage{times}
\usepackage{latexsym}
\usepackage{booktabs}
\usepackage{tabularx}
\usepackage{colortbl}

\usepackage[T1]{fontenc}

\usepackage[utf8]{inputenc}
\usepackage{microtype}
\usepackage{caption}
\usepackage{subcaption}
\usepackage{adjustbox}
\usepackage{multirow}
\usepackage{soul}
\usepackage{amsfonts}
\usepackage{amsmath}
\usepackage{bbding}

\usepackage{microtype}

\title{
Chaining Simultaneous Thoughts for Numerical Reasoning
}

\author{
    Zhihong Shao, Fei Huang, Minlie Huang\thanks{*Corresponding author: Minlie Huang.}\\
    The CoAI group, DCST, Tsinghua University, Institute for Artificial Intelligence; \\
    State Key Lab of Intelligent Technology and Systems; \\
    Beijing National Research Center for Information Science and Technology; \\
    Tsinghua University, Beijing 100084, China \\
    {\tt \{szh19, f-huang18\}@mails.tsinghua.edu.cn} \\
    {\tt aihuang@tsinghua.edu.cn}}

\begin{document}
\maketitle
\begin{abstract}
Given that rich information is hidden behind ubiquitous numbers in text, numerical reasoning over text should be an essential skill of AI systems.
To derive precise equations to solve numerical reasoning problems, previous work focused on modeling the structures of equations, and has proposed various structured decoders.
Though structure modeling proves to be effective, these structured decoders construct a single equation in a pre-defined autoregressive order,
potentially placing an unnecessary restriction on how a model should grasp the reasoning process.
Intuitively, humans may have numerous pieces of thoughts popping up in no pre-defined order; thoughts are not limited to the problem at hand, and can even be concerned with other related problems. By comparing diverse thoughts and chaining relevant pieces, humans are less prone to errors.
In this paper, we take this inspiration and propose CANTOR, a numerical reasoner that models reasoning steps using a directed acyclic graph where we produce diverse reasoning steps simultaneously without pre-defined decoding dependencies, and compare and chain relevant ones to reach a solution.
Extensive experiments demonstrated the effectiveness of CANTOR under both fully-supervised and weakly-supervised settings.

\end{abstract}

\section{Introduction}
Numerical reasoning over text is an essential skill for a neural model to help analyze rich numerical information from large-scale textual data \cite{DBLP:conf/emnlp/ChenCSSBLMBHRW21}.
Many question answering benchmarks \cite{DBLP:conf/naacl/DuaWDSS019,DBLP:conf/naacl/PatelBG21} have been created to promote the numerical reasoning ability of neural models, where, typically, models are required to answer questions about given contexts with numerical answers.
This is challenging, as it requires comprehensive structural analyses of text as well as precise and possibly complex deduction.
\begin{figure}[t!]
    \centering
    \includegraphics[width=0.49\textwidth]{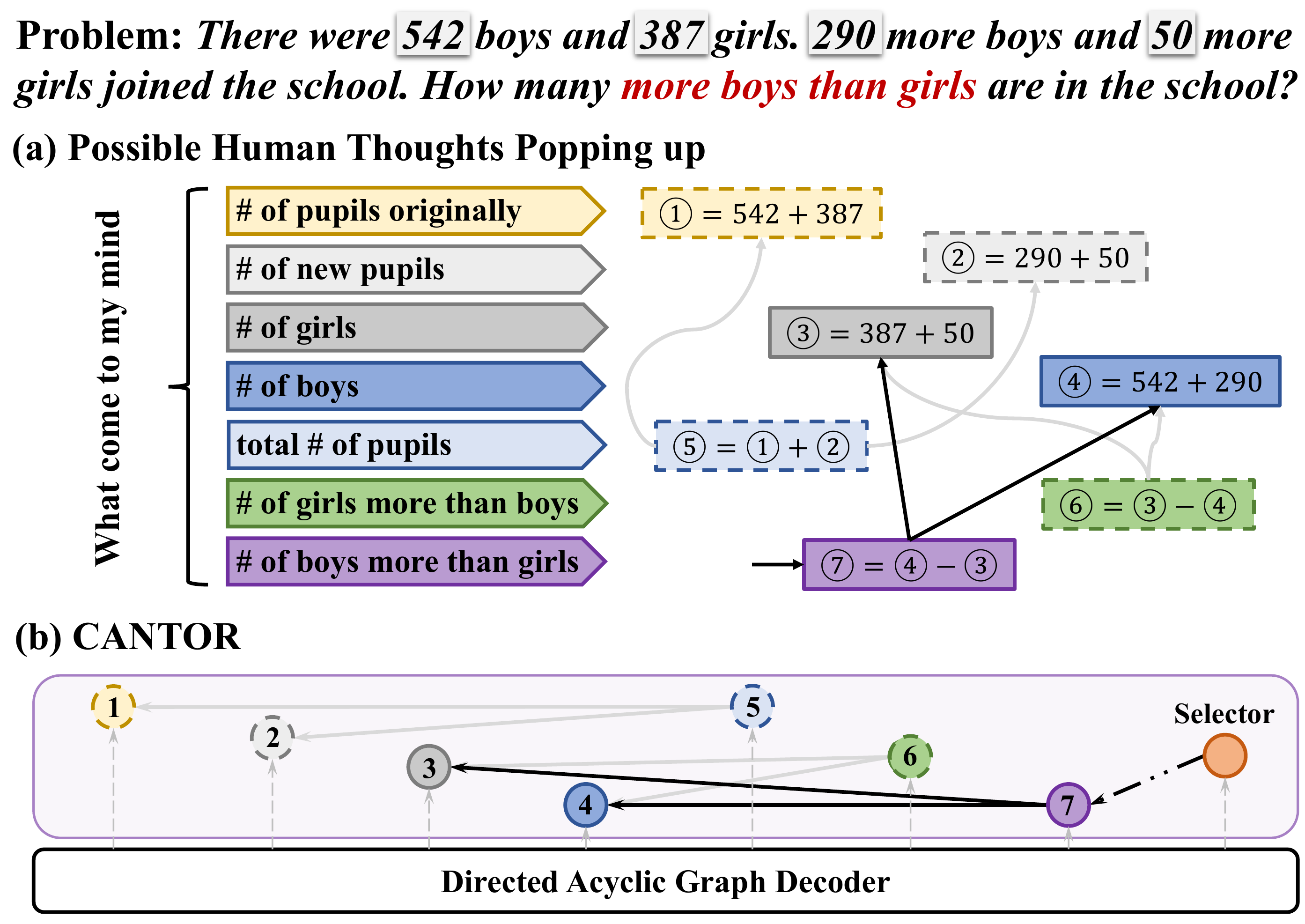}
    \caption{(a) Possible pieces of human thoughts that pops up in no pre-defined order; (b) How our model captures the reasoning process similarly. Reasoning steps inside solid frames and dashed frames are necessary and loosely-relevant ones, respectively.}
    \label{fig:example}
\end{figure}

Existing models mostly decode the equations and return the execution results. To better exploit structures of equations, many complex structured decoders \cite{DBLP:conf/ijcai/XieS19,DBLP:conf/aaai/CaoH0021} have been proposed and significantly outperform sequential decoding \cite{DBLP:journals/corr/abs-2105-08928}. However, all these methods construct a single equation in a pre-defined order (e.g., top-down or bottom-up order), which may place an unnecessary restriction on how a model should grasp the reasoning process.

Intuitively, after reading a reasoning problem, humans may have several pieces of thoughts which pop up in no pre-defined order,
and finalize a solution by comparing and chaining relevant pieces.
Take Fig \ref{fig:example}(a) for example.
Possible thoughts include necessary reasoning steps (e.g., how to get the number of boys and girls separately) and loosely-relevant ones (e.g., those learned from previous similar questions like ``how many \textcolor{red}{more girls than boys} are in the school?'').
There is arguably no pre-defined strict order where a thought should conditionally emerge after some other thoughts.
By comparing these diverse thoughts, we finally select and chain proper ones to reach a solid solution, which will be less prone to mistakes.

In this paper, we propose \textbf{CANTOR}, which compares and \textbf{C}hains simult\textbf{AN}eous \textbf{T}h\textbf{O}ughts for numerical \textbf{R}easoning.
As in Fig \ref{fig:example}(b), CANTOR constructs a Directed Acyclic Graph (DAG) of diverse reasoning steps in a non-autoregressive way: all vertices are produced simultaneously, which correspond to operations like addition, and edges in the graph are constructed by chaining operations with their best-matched operands; the final equation is a selected sub-graph in the whole DAG.
With no pre-defined decoding order, logical dependencies among reasoning steps are freely captured by the model internally.
With our training methods, CANTOR captures diverse reasoning steps at different vertices, and learns to prune away possibly-distracting candidates during both training and inference, resulting in chaining reasoning steps that are more consistent with given problems.


To summarize, compared with previous models with structured decoding, CANTOR has no pre-defined restrictions on the decoding dependencies
while also benefiting from modeling the structures of equations.
Besides, by comparing diverse reasoning steps and chaining logically consistent ones, our model is less prone to errors.
Our model establishes a new state-of-the-art record on two math word problem datasets under the fully-supervised setting, and is also applicable to weakly-supervised scenarios (where problems are only annotated with final answers, and the equations are unavailable) with significant improvements over baselines.
Though not directly comparable, on two numerical reasoning datasets, fully-supervised CANTOR achieves even higher accuracies than hundreds of times larger language models (e.g., PaLM-62B \cite{DBLP:journals/corr/abs-2204-02311}) that use the effective chain-of-thought prompting technique \cite{DBLP:journals/corr/abs-2201-11903}, demonstrating CANTOR's great potential.

\section{Related Work}
\noindent
\textbf{Numerical Reasoning}
Numerical reasoning tasks can be formulated in many ways \cite{DBLP:conf/acl/MishraMVSCBK22}, such as (1) question answering with numerical answers directly derived from arithmetic operations \cite{DBLP:conf/naacl/Koncel-Kedziorski16,DBLP:conf/emnlp/WangLS17,DBLP:conf/naacl/DuaWDSS019,DBLP:conf/naacl/AminiGLKCH19,DBLP:conf/acl/MiaoLS20,DBLP:conf/naacl/PatelBG21}, (2) or other tasks like quantitative natural language inference \cite{DBLP:conf/conll/RavichanderNRH19} whose expected outputs are non-numerical but require implicit arithmetic reasoning.
In this work, we focus on the former type of task which is widely studied.
To generate equations precisely, previous work proposed to enhance number-related representations in problem encoding \cite{DBLP:conf/acl/ZhangWLBWSL20,DBLP:conf/coling/ShenJ20,DBLP:journals/corr/abs-2107-13435}, re-rank equation samples with a verifier \cite{DBLP:conf/emnlp/ShenYLSJ0021,DBLP:journals/corr/abs-2110-14168}, or exploit the structures of equations with complex top-down tree-structured decoding \cite{DBLP:conf/ijcai/XieS19,DBLP:conf/acl/LiZYZLLC22} or bottom-up DAG-structured decoding \cite{DBLP:conf/aaai/CaoH0021,DBLP:conf/acl/JieLL22}.
Our numerical reasoner also models equations with DAGs but with three major differences:
(1) there is no pre-defined decoding order which may place unnecessary burden on how a model should learn the dependencies among operations;
(2) the decoding process is largely simplified, which is reduced to simultaneous predictions of an operator and operands at each vertex of a graph;
(3) our model explores diverse operations in a DAG and is trained to compare and chain relevant ones, so that logical consistency between given problems and equations are better captured during both training and inference.

\noindent
\textbf{Non-Autoregressive Decoding}
Our model is also relevant to non-autoregressive decoding.
For machine translation, non-autoregressive translation \cite{DBLP:conf/iclr/Gu0XLS18,DBLP:conf/icml/GhazvininejadKZ20,DBLP:conf/icml/DuTJ21} aims at fast inference; the recently proposed DA-Transformer \cite{DBLP:journals/corr/abs-2205-07459}, which utilizes a DAG to capture diverse translations, has made great progress in bridging the performance gap with autoregressive models.
Recent work has also proposed non-autoregressive models for efficient task-oriented semantic parsing \cite{DBLP:conf/naacl/BabuSAAFG21,DBLP:conf/emnlp/ShrivastavaCBDA21}, which achieved comparable performance with autoregressive parsers.
All these methods model a target as a sequence and adopt token-wise decoding (one token at a position). By contrast, we model a target as a DAG and adopt step-wise decoding (one complete reasoning step at each vertex), which facilitates structure modeling and learning meaningful vertex representations.
Experimental results show that our model significantly outperforms both autoregressive and non-autoregressive baselines.
Notably, for open text generation, autoregressive methods are probably still the better choice for strong probabilistic modeling of diverse targets.
However, for the numerical reasoning task we focus on, it is the logical relationships among quantities (both known and unknown in a given problem) that matter, and non-autoregressive methods, with proper designs, suffice to decode equations precisely and can provide new perspectives on how numerical reasoning can be better grasped by neural models.

\section{Task Definition}
Given a problem description $X$ which mentions a list of numbers $\mathcal{N}=\{n_1, n_2, ..., n_{|\mathcal{N}|}\}$, our task is to return the numerical answer $A$ which is derived from an equation $Y$ that takes arithmetic operations (e.g., addition, subtraction, multiplication, division, and exponentiation) on $\mathcal{N}$ as well as a set of pre-defined constants $\mathcal{C}=\{c_1, c_2, ..., c_{|\mathcal{C}|}\}$.

For scenarios that consider only binary operators\footnote{In this paper, we only consider binary operators while it is feasible to extend our model to utilize other n-ary operators.}, a ground-truth equation $Y$ can be formally defined as follows:
\begin{equation}
    \small
    \begin{split}
        Y =& \{y_1, y_2, ..., y_{|Y|}\}, \ y_i = \langle y_i^f, y_i^a, y_i^b \rangle \\
        s.t.\ \ &y_i^f \in \mathcal{F} \wedge y_i^a, y_i^b \in \mathcal{C} \cup \mathcal{N} \cup \{y_k|k<i\}
    \end{split}\nonumber
\end{equation}
where $\mathcal{F}$ is the set of pre-defined operators. $y_i$ is an operation that applies the operator $y_i^f$ to the two operands $y_i^a$ and $y_i^b$. $Y$ can be directly transformed into a DAG with $c_i$, $n_i$, and $y_i$ being vertices, and $y_i \rightarrow y_i^a$ and $y_i \rightarrow y_i^b$ being edges. The final operation (the root vertex) $y_{|Y|}$ returns the answer.

\begin{figure*}[t!]
    \centering
    \includegraphics[width=0.95\textwidth]{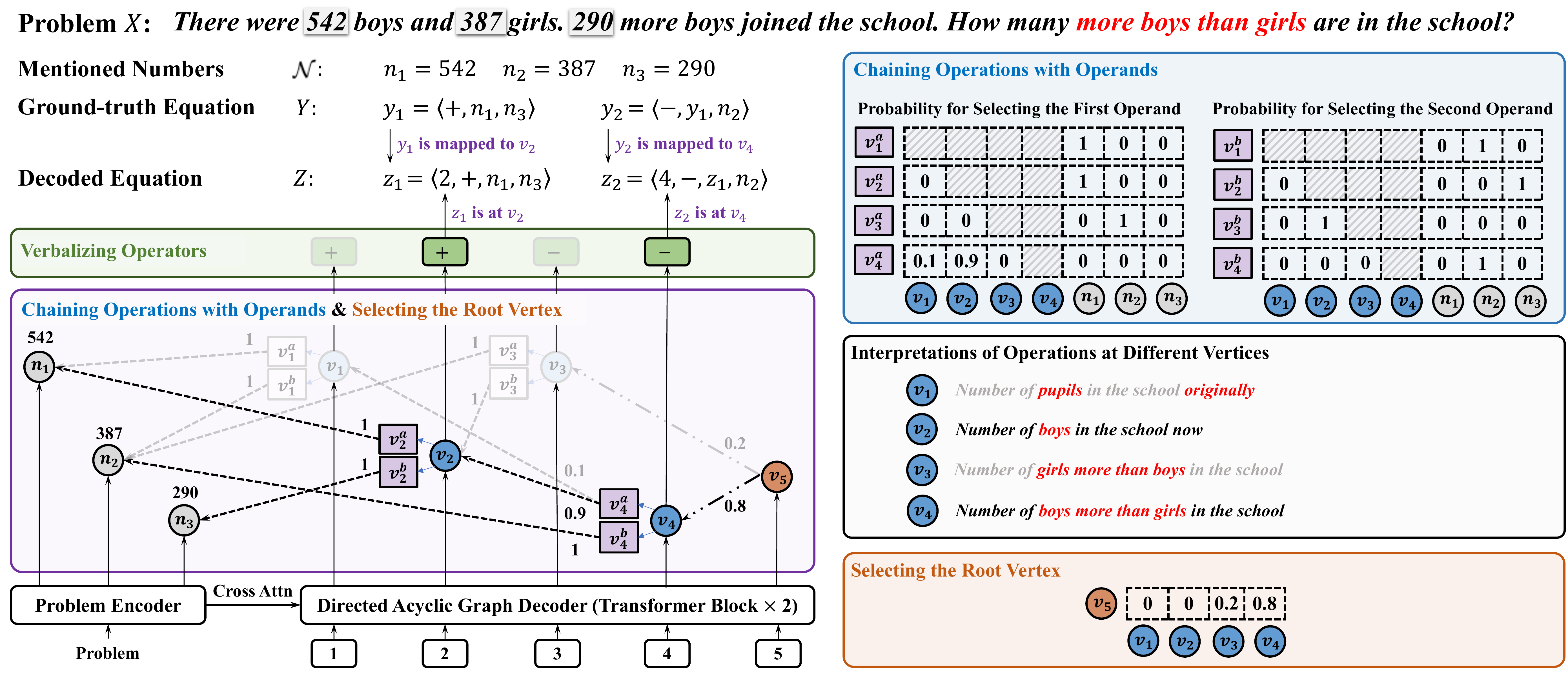}
    \caption{Overview of CANTOR. CANTOR models diverse operations using a DAG. Each vertex corresponds to an operation, which is chained with its operands via edges in the graph. We decode an equation by simultaneously verbalizing operators at each vertex, chaining operations with operands, and selecting the root vertex; the selected root vertex along with all its descendants is the resulting equation in a DAG format. In this example, the ground-truth equation $Y$ can be represented by the decoded sub-graph $Z$, as mapping $y_1$ to $v_2$ and $y_2$ to $v_4$ produces $Z$ exactly.}
    \label{fig:architecture}
\end{figure*}

\section{CANTOR}
\subsection{Overview}
We propose to model diverse reasoning steps with a DAG.
Vertices of the graph correspond to reasoning steps which are decoded in parallel.
This is analogous to humans' burst of thoughts after reading a reasoning problem.
No pre-defined restriction is placed on how a reasoning step should conditionally depend on others; logical dependencies among reasoning steps are captured by the model internally.
Our DAG also allows the model to explore diverse reasoning steps at different vertices, including necessary or wrong ones;
the model is trained to compare the semantics of diverse reasoning steps and chain the most proper ones to be the final equation, which benefits model performance.

\subsection{Architecture}
Our model (Fig \ref{fig:architecture}) comprises a pre-trained Transformer encoder (e.g., RoBERTa) and a shallow Transformer-based DAG decoder.
The encoder encodes a problem $X$; from the encoder outputs, we can obtain the representations of mentioned numbers $\mathbf{N}=[\mathbf{n_1}, \mathbf{n_2}, ..., \mathbf{n_{|\mathcal{N}|}}] \in \mathbb{R}^{d \times |\mathcal{N}|}$ ($d$ is the hidden size).
The DAG decoder, with positional embeddings as inputs and cross attention over encoder outputs, produces representations for $L$ vertices $\mathcal{V} = \{v_1, v_2, ..., v_L\}$ in a non-autoregressive way, which are denoted as $\mathbf{V} = [\mathbf{v_1}, \mathbf{v_2}, ..., \mathbf{v_L}] \in \mathbb{R}^{d \times L}$.
Each vertex representation encodes the semantics of a reasoning step, including its operator, the expected operands, and the meaning of the resulting quantity.
We then verbalize the operator for each vertex and chain it with its best-matched operands in parallel, and finally, select one root vertex and return its execution result.
The selected root vertex along with its vertex descendants constitutes a decoded sub-graph, which is also the DAG representation of an equation.
Let $Z$ be the decoded sub-graph, which can be formulated as:
\begin{equation}
    \small
    \begin{split}
        Z =& \{z_1, z_2, ..., z_{|Z|}\},\ z_j = \langle p_j, z_j^f, z_j^a, z_j^b \rangle \\
        s.t.\ \ & 1 \le p_1 < p_2 < ... < p_{|Z|} \le L \\
        & z_j^f \in \mathcal{F} \wedge z_j^a, z_j^b \in \mathcal{C} \cup \mathcal{N} \cup \{z_k|k < j\}
    \end{split}\nonumber
\end{equation}
where $z_j$ is the operation for the vertex at position $p_j$, with $z_j^f$ being the operator, and $z_j^a$ and $z_j^b$ being its operands.
$p_{|Z|}$ is the index of the root vertex, and $\{p_j|j<|Z|\}$ are indices of its vertex descendants.

The probability of a target equation $Y$ can be formulated as follows:
\begin{equation}
    \small
    \begin{split}
        P_\mathbf{\theta}(Y|X) = \sum_{Z} P_\mathbf{\theta}(Y|Z,X) P_\mathbf{\theta}(Z|X)
    \end{split}
    \label{eq:latent_variable_model}
\end{equation}

\noindent
\textbf{Definition of $P_\mathbf{\theta}(Y|Z, X)$:}
Given $Y$ and $Z$, $P_\mathbf{\theta}(Y|Z, X)$ is defined to be 1 if and only if
mapping $y_i$ to the vertex at position $p_i$ ($\forall 1 \le i \le |Y|$) produces $Z$ exactly;
otherwise, $P_\mathbf{\theta}(Y|Z, X)$ is 0.

Therefore, $P_\theta(Y|X)$ can be re-written as:
\begin{equation}
    \small
    \begin{split}
        P_\theta(Y|X) &= \sum_{Z \in \Gamma} P_\theta(Z|X) \\
        \Gamma &= \{Z|P_\theta(Y|Z,X)=1\}
    \end{split}
\end{equation}
Any $Z \in \Gamma$ is a DAG representation of $Y$ (see the example in Fig \ref{fig:architecture}).
For a given $Y$, $\Gamma$ can be created by enumerating $\{p_1, ..., p_{|Y|}\}$ that satisfies $1 \le p_1 < ... < p_{|Y|} \le L$, and then mapping $y_i$ to $v_{p_i}$ ($\forall 1 \le i \le |Y|$).
Therefore, $|\Gamma| = \tbinom{L}{|Y|}$.

\noindent
\textbf{Definition of $P_\mathbf{\theta}(Z|X)$:}
$P_\mathbf{\theta}(Z|X)$ can be further decomposed based on operations in $Z$:
{\small
\begin{align}
    &P_\theta(Z|X) = P_r(p_{|Z|}|X) \prod_{j=1}^{|Z|} P_z(z_j|p_j, X) \label{eq:decomp} \\
    &P_z(z_j|p_j, X) = P_f(z_j^f|p_j, X) P_a(z_j^a|p_j, X) P_b(z_j^b|p_j, X) \nonumber
\end{align}
}%
where $P_r(\cdot)$ and $P_z(\cdot)$ are the probability functions of the root vertex and an operation, respectively;
$P_f(\cdot)$ and $P_a(\cdot)$ ($P_b(\cdot)$) are for operator verbalization and operand matching, respectively.

\subsubsection{Verbalizing Operators}
We verbalize an operator for each vertex based on its representation:
\begin{equation}
    \small
    \begin{split}
        P_f(z_j^f|p_j, X) = \mbox{softmax}(\mathbf{W_f} \mathbf{v_{p_j}})
    \end{split} \nonumber
\end{equation}
where $\mathbf{W_f} \in \mathbb{R}^{|\mathcal{F}|\times d}$ is trainable parameters.

\subsubsection{Chaining Operations with Operands}
Each operation is connected with its best-matched operands chosen from all available quantities (including the other operations, constants, and mentioned numbers).
Let $\mathbf{C} = [\mathbf{c_1}, \mathbf{c_2}, ..., \mathbf{c_{|C|}}]^\top$ be the embedding matrix for pre-defined constants.
Then the representation matrix for all quantities is denoted as $\mathbf{Q} = [\mathbf{V},\mathbf{C},\mathbf{N}] \in \mathbb{R}^{d \times (L + |\mathcal{C}| + |\mathcal{N}|)}$.
The probability distribution over candidates when predicting the first operand for the vertex at position $p_j$ can be computed as follows:
\begin{equation}
    \small
    \begin{split}
        P_a(z_j^a|p_j, X) = \mbox{softmax}(\frac{(\mathbf{W_q} \mathbf{Q})^\top \mathbf{v_{p_j}^a}}{\sqrt{d}}),\ \mathbf{v_{p_j}^a} = \mathbf{W_a} \mathbf{v_{p_j}}
    \end{split}\nonumber
\end{equation}
The probability for predicting the second operand $P_b(\cdot)$ can be computed likewise.
$\mathbf{W_q}, \mathbf{W_a}, \mathbf{W_b}  \in \mathbb{R}^{d \times d}$ are trainable parameters.
To avoid cycles in the graph, we apply probability masking so that a vertex can not use itself or vertices with larger indices as its operands.

\subsubsection{Selecting the Root Vertex and Finalizing the Equation}
The final equation is represented by a sub-graph of the whole DAG, which comprises a selected root vertex and all its descendants.
We introduce a special vertex $v_{L+1}$ at position $L+1$ of the decoder, and use its representation to select the best-matched root vertex:
\begin{equation}
    \small
    \begin{split}
        P_r(p_{|Z|}|X) = \mbox{softmax}(\frac{(\mathbf{W_q}\mathbf{V})^\top \mathbf{v_{L+1}}}{\sqrt{d}})
    \end{split}\nonumber
\end{equation}
where $\mathbf{v_{L+1}}$ is the representation of $v_{L+1}$, computed the same way as other vertex representations.

\subsection{Training}
To capture diverse reasoning steps at different vertices, we explore four training methods\footnote{See Appendix \ref{appendix:training} for implementation details of hard EM and MML.}, namely, na\"ive mapping, hard EM, MML, and hard EM with annealing.
Notably, as will be discussed by Section \ref{sec:ablate_training_methods}, in practice, one has no need to consider all four training methods; hard EM with annealing should be the default choice.

\subsubsection{Na\"ive Mapping}
A na\"ive way of mapping from $Y$ to $\mathcal{V}$ is to map $y_i$ to $v_i$ ($\forall 1 \le i \le |Y|$). Let $Z^\prime$ be the resulting sub-graph, then the training objective is:
\begin{equation}
    \small
    \begin{split}
        \mathcal{L} = -\log P_\theta(Z^\prime|X)
    \end{split}
\end{equation}
which leaves $\{v_j||Y| < j \le L\}$ unused.

\subsubsection{Hard EM}
Hard EM is to optimize the probability of $Z^*$ that best aligns with $Y$:
\begin{equation}
    \small
    \begin{split}
        \mathcal{L} = -\log P_\theta(Z^*|X),\ Z^* = arg\max_{Z \in \Gamma} P_\theta(Z|X)
    \end{split}
\end{equation}
As $|\Gamma|$ can be quite large\footnote{Suppose $L=60$ and $|Y|=15$, then $|\Gamma| \approx 5\times10^{13}$.},
we use beam search to find $Z^*$ approximately, which is feasible as $P_\theta(Z|X)$ can be factorized into probabilities of constituent operations of $Z$ (Eq \ref{eq:decomp}).
Notably, the probability of an operation depends on which vertices its operands (if being operations) are mapped to.
We search $Z$ by iteratively determining where to map $y_i$, until $y_{|Y|}$ is settled.


\subsubsection{MML}
MML optimizes the marginal likelihood of $Z$:
\begin{equation}
    \small
    \begin{split}
        \mathcal{L} = - \log \sum_{Z \in \Gamma} P_\theta(Z|X)
    \end{split}
\end{equation}
Marginalization is expensive due to the large size of $\Gamma$.
We therefore adopt a strong (but risky) assumption so that we can use dynamic programming to marginalize $P_\theta(Z|X)$ in polynomial time.
Specifically, for any operation $y_i$, we assume that the two sub-graphs rooted at $y_i^a$ and $y_i^b$ respectively (in the DAG counterpart of $Y$) are independently mapped to $\{v_1, v_2, ..., v_L\}$.
Notably, with this assumption, we in fact marginalize $P_\theta(Z|X)$ over a superset of $\Gamma$ and even allow mapping multiple operations to a single vertex.
However, we empirically found that MML (with this assumption) works well on short equations\footnote{We provide detailed discussion on the properties of our MML in Appendix \ref{appendix:limit_mml}.}, and can be used to warm up hard EM.

\subsubsection{Hard EM with Annealing}
To avoid optimizing the model on its early decisions, we follow \citet{DBLP:conf/emnlp/MinCHZ19} to apply annealing to hard EM: we optimize the model using MML for $\tau$ training steps and use hard EM afterwards.

\subsection{Inference}
During inference, we adopt greedy decoding which conducts the argmax operation for operator prediction, operand matching, and root vertex selection in parallel. The execution result at the root vertex is returned as the numerical answer.

\section{Experiments}
\subsection{Datasets}
\begingroup
\setlength{\tabcolsep}{3pt} 
\renewcommand{\arraystretch}{1} 
\begin{table}[htb]
    \centering
	\adjustbox{max width=.47\textwidth}{
    \begin{tabular}{lcccccc}
        \hline
        Dataset & Train & Dev & Test & $|\mathcal{C}|$ & $\mathcal{F}$ & $\max |Y|$ \\
	    \hline
	    \multicolumn{7}{l}{\textit{Fully-Supervised MWP Solving}} \\
	    MathQA & 16,191 & 2,411 & 1,605 & 24 & $\{+,-,\times,/,**\}$ & 15 \\
	    SVAMP & 3,138 & - & 1,000 & 17 & $\{+,-,\times,/\}$ & 7 \\
	    \hline
	    \multicolumn{7}{l}{\textit{Weakly-Supervised Discrete Reasoning}} \\
	    DROP\textsubscript{num} & 46,973 & 5,850 & - & 2 & $\{+,-\}$ & 1 \\
	    DROP & 77,409 & 9,536 & 9,615 & 2 & $\{+,-\}$ & 1 \\
	    \hline
    \end{tabular}
    }
    \caption{Data statistics. Note that DROP\textsubscript{num} and DROP are only annotated with answer texts but not equations $Y$; we followed previous work to enumerate binary operations that evaluate to the answers ($\max |Y| = 1$).}
    \label{tab:data}
\end{table}
\endgroup

\noindent
We applied CANTOR to Math Word Problem (MWP) solving under the fully-supervised setting and discrete reasoning under a weakly-supervised setting.
Data statistics are shown in Table \ref{tab:data}.

\noindent
\textbf{Fully-Supervised MWP Solving}

\noindent
\textbf{(a) MathQA} \cite{DBLP:conf/naacl/AminiGLKCH19} consists of GRE level math problems from multiple domains. We used the dataset from \citet{DBLP:conf/acl/JieLL22} which has wrongly-annotated instances removed.

\noindent
\textbf{(b) SVAMP} \cite{DBLP:conf/naacl/PatelBG21} was created for robustness evaluation, which consists of problems from the ASDiv-A dataset \cite{DBLP:conf/acl/MiaoLS20} with manual perturbations. We strictly followed \citet{DBLP:conf/naacl/PatelBG21} to use both MAWPS \cite{DBLP:conf/naacl/Koncel-Kedziorski16} and ASDiv-A for training and SVAMP for testing.

\noindent
\textbf{Weakly-Supervised Discrete Reasoning}

\noindent
\textbf{(a) DROP\textsubscript{num}} \cite{DBLP:conf/naacl/DuaWDSS019} consists of all problems with numerical answers from the reading comprehension dataset called DROP. Problems are only annotated with final answers but not the corresponding equations.

\noindent
\textbf{(b) DROP} is a reading comprehension dataset consisting of problems with different types of answers, e.g., number, date, and span(s).

\subsection{Metrics}
For MWP solving, we evaluated models with \textbf{value accuracy} and \textbf{equation accuracy}. Previous work evaluated equation accuracy with string matching, failing to recognize positive equations that are structurally different from the ground-truth. In our evaluation, an equation is considered correct if it has consistent results with the annotated equation for 100 random replacements of numbers mentioned in the problem. For discrete reasoning on DROP, we followed previous work to use \textbf{F1}.

\subsection{Baselines}
We considered the following three categories:

\noindent
\textbf{Sequential Models} generate an equation sequentially based on a given problem.
mBERT2Seq \cite{DBLP:journals/corr/abs-2105-08928} comprises a multilingual BERT \cite{DBLP:conf/naacl/DevlinCLT19} encoder and an LSTM \cite{DBLP:journals/neco/HochreiterS97} decoder.
We also compared our model with two sequential models from \cite{DBLP:journals/corr/abs-2109-00799}, namely, GPT-2 \cite{Radford2019LanguageMA} and RoBERTaGen \cite{DBLP:journals/corr/abs-1907-11692}.

\noindent
\textbf{Structured Models} utilize structured autoregressive decoders to generate an equation.
Graph2Tree \cite{DBLP:conf/acl/ZhangWLBWSL20} and D\textsc{educt}R\textsc{easoner} \cite{DBLP:conf/acl/JieLL22} are the representative tree-structured model and DAG-structured model, respectively.

\noindent
\textbf{Tagging-based Models} refer to the arithmetic modules of those modular networks dominant on DROP, which assign a plus, minus, or zero to each constant and number mentioned in a problem, and return the sum of the signed numbers.
TASE \cite{DBLP:conf/emnlp/SegalESGB20} is a representative modular network which consists of modules specialized for different types of answers, e.g., a tagging-based arithmetic module, a count module, and modules for span-typed answers.
We referred to a TASE model with only an arithmetic module as TASE\textsubscript{arith}.

\begingroup
\setlength{\tabcolsep}{3pt} 
\renewcommand{\arraystretch}{1} 
\begin{table}[t!]
    \centering
	\adjustbox{max width=.42\textwidth}{
    \begin{tabular}{lcc}
        \hline
        Model & Dev & Test \\
        \hline
        \multicolumn{3}{l}{\textit{Sequential Model}}\\
        mBERT2Seq \cite{DBLP:journals/corr/abs-2105-08928} & - & 77.1 \\
        \hline
        \multicolumn{3}{l}{\textit{Structured Model}} \\
        Graph2Tree \cite{DBLP:conf/acl/ZhangWLBWSL20} & - & 69.5 \\
        BERT2Tree \cite{DBLP:conf/acl/LiZYZLLC22} & - & 73.8 \\
        D\textsc{educt}R\textsc{easoner} \cite{DBLP:conf/acl/JieLL22} & - & 78.6 \\
	    \hline
	    CANTOR & \textbf{81.7} & \textbf{82.9} \\
	    \hline
    \end{tabular}
    }
    \caption{Value accuracy on MathQA.}
    \label{tab:mathqa}
\end{table}
\endgroup

\subsection{Implementation Details}
For all experiments, we used two Transformer blocks \cite{DBLP:conf/nips/VaswaniSPUJGKP17} as the DAG decoder, which was trained with random initialization.

For MWP solving, we used RoBERTa\textsubscript{base} as the problem encoder.
We experimented with different training methods whose effect on model performance will be discussed in Section \ref{sec:ablate_training_methods} with the graph size $L$ set to 60 and the beam size $B$ for hard EM set to 20. We further investigated the effect of graph size $L$ (Table \ref{tab:ablate_graph_size} in Section \ref{sec:ablate_graph_size}) and beam size $B$ (Table \ref{tab:ablate_beam_size} in Appendix \ref{sec:ablate_beam_size}).
The best model on MathQA used hard EM with annealing ($\tau = 2,000, B=20$), with $L=80$, and the best model on SVAMP used MML, with $L=60$.
Following previous work, all experiments on SVAMP were run with 5 random seeds, with both the average performance and standard deviation reported.

For discrete reasoning on DROP, we followed TASE to use RoBERTa\textsubscript{large} for encoding and MML for training. $L$ was chosen from $\{5, 10\}$ based on F1.
The best models on DROP\textsubscript{num} and DROP used $L=5$ and $L=10$, respectively.

\begingroup
\setlength{\tabcolsep}{3pt} 
\renewcommand{\arraystretch}{1} 
\begin{table}[t!]
    \centering
	\adjustbox{max width=.42\textwidth}{
    \begin{tabular}{lc}
        \hline
        Model & Test \\
        \hline
        \multicolumn{2}{l}{\textit{Sequential Model}}\\
        GPT-2 \cite{DBLP:journals/corr/abs-2109-00799} & 25.7 \\
        RoBERTaGen \cite{DBLP:journals/corr/abs-2109-00799} & 30.3 \\
        \hline
        \multicolumn{2}{l}{\textit{Structured Model}} \\
        RoBERTa-Graph2Tree \cite{DBLP:conf/naacl/PatelBG21} & 43.8 \\
        BERT2Tree \cite{DBLP:conf/acl/LiZYZLLC22} & 32.4 \\
        D\textsc{educt}R\textsc{easoner} \cite{DBLP:conf/acl/JieLL22} & 45.1 \\
	    \hline
	    CANTOR & \textbf{49.6}\textsubscript{$\pm 0.63$} \\
	    \hline
    \end{tabular}
    }
    \caption{Value accuracy on SVAMP.}
    \label{tab:svamp}
\end{table}
\endgroup

\begingroup
\setlength{\tabcolsep}{3pt} 
\renewcommand{\arraystretch}{1} 
\begin{table}[t!]
    \centering
	\adjustbox{max width=.45\textwidth}{
    \begin{tabular}{ccccccccc}
        \hline
        \multirow{3}{*}{Breakdown} & \multicolumn{4}{c}{MathQA} & \multicolumn{4}{c}{SVAMP} \\
        \cmidrule(lr){2-5} \cmidrule(lr){6-9}
        & \multicolumn{2}{c}{Baseline} & \multicolumn{2}{c}{CANTOR} & \multicolumn{2}{c}{Baseline} & \multicolumn{2}{c}{CANTOR} \\
        \cmidrule(lr){2-3} \cmidrule(lr){4-5} \cmidrule(lr){6-7} \cmidrule(lr){8-9}
        & Equ. & Val. & Equ. & Val. & Equ. & Val. & Equ. & Val. \\
        \hline
        \multicolumn{9}{l}{\textit{Breakdown w.r.t. \# Operation}} \\
        1 & 76.4 & 79.1 & \textbf{78.2} & \textbf{80.0} & 48.8 & 49.1 & \textbf{54.9} & \textbf{55.2} \\
        2 & 81.0 & 83.5 & \textbf{83.1} & \textbf{84.8} & \textbf{31.2} & \textbf{32.1} & 30.7 & 31.6 \\
        3 & 80.8 & 83.6 & \textbf{82.6} & \textbf{86.7} & - & - & - & - \\
        4 & 78.5 & 82.0 & \textbf{81.3} & \textbf{84.4} & - & - & - & - \\
        $\ge 5$ & 65.7 & 71.3 & \textbf{74.4} & \textbf{79.4} & - & - & - & - \\
        \hline
        \multicolumn{9}{l}{\textit{Breakdown w.r.t. Equation Novelty}} \\
        Seen & 90.5 & 91.4 & \textbf{95.2} & \textbf{96.1} & 48.8 & 49.2 & \textbf{53.5} & \textbf{53.8} \\
        Unseen & 34.5 & 45.8 & \textbf{38.3} & \textbf{49.3} & 12.2 & 13.9 & \textbf{15.8} & \textbf{16.9} \\
        \hline
        \multicolumn{9}{l}{\textit{Overall Performance}} \\
        Full & 74.7 & 78.6 & \textbf{79.2} & \textbf{82.9} & 44.6 & 45.1 & \textbf{49.2} & \textbf{49.6} \\
        \hline
    \end{tabular}
    }
    \caption{Breakdowns of performance on the MWP solving task. \textit{Baseline} refers to the previous best model D\textsc{educt}R\textsc{easoner}. \textit{Equ.} and \textit{Val.} are equation accuracy and value accuracy, respectively.}
    \label{tab:mwp_breakdown}
\end{table}
\endgroup

\begingroup
\setlength{\tabcolsep}{3pt} 
\renewcommand{\arraystretch}{1} 
\begin{table}[t!]
    \centering
	\adjustbox{max width=.35\textwidth}{
    \begin{tabular}{ccccc}
        \hline
        \multirow{2}{*}{Variations} & \multicolumn{2}{c}{Baseline} & \multicolumn{2}{c}{CANTOR} \\
        \cmidrule(lr){2-3} \cmidrule(lr){4-5} 
        & Equ. & Val. & Equ. & Val. \\
        \hline
        Question Sensitivity & 21.6 & 22.3 & \textbf{29.6} & \textbf{30.3} \\
        Reasoning Ability & 49.5 & 49.8 & \textbf{53.2} & \textbf{53.4} \\
        Structural Invariance & 37.3 & 38.1 & \textbf{42.4} & \textbf{43.2} \\
        \hline
    \end{tabular}
    }
    \caption{A breakdown of robustness evaluation w.r.t. different variations in SVAMP. \textit{Baseline} refers to the previous best model D\textsc{educt}R\textsc{easoner}. \textit{Equ.} and \textit{Val.} are equation accuracy and value accuracy, respectively.}
    \label{tab:svamp_breakdown_variations}
\end{table}
\endgroup
\subsection{Results for MWP Solving}
As shown by Table \ref{tab:mathqa} and Table \ref{tab:svamp}, CANTOR established a new state-of-the-art record on MathQA and SVAMP
with large improvements.
The fine-grained analyses in Table \ref{tab:mwp_breakdown} and Table \ref{tab:svamp_breakdown_variations} show that
CANTOR (1) outperforms the best baseline on nearly all problems of different levels of complexity measured by the number of operations needed, (2) is better at exploiting equation templates\footnote{Equation templates are equations with numbers replaced with placeholders, e.g., \texttt{const\_10 + num@7} adds 10 to the 7-th number in a problem.} seen in training or creating novel ones to solve problems, (3) and is more robust to different types of variations, including those that evaluate question sensitivity (whether questions asked in problems are ignored in prediction), reasoning ability (how predictions are adjusted to subtle changes in given problems), and structural invariance (whether predictions are invariant to structural changes of given problems that preserve the reasoning logic).



\subsection{Results for Discrete Reasoning}
\begingroup
\setlength{\tabcolsep}{3pt} 
\renewcommand{\arraystretch}{1} 
\begin{table}[htb]
    \centering
    \subfloat[DROP\textsubscript{num}]{
	\adjustbox{max width=.13\textwidth}{
    \begin{tabular}{cc}
        \hline
         Model & Dev \\
        \hline
        TASE\textsubscript{arith} & 76.4 \\
        CANTOR & \textbf{78.1}\\
        \hline
    \end{tabular}
    }
    \label{tab:drop_num}
    }
    \subfloat[DROP]{
	\adjustbox{max width=.32\textwidth}{
    \begin{tabular}{cccc}
        \hline
         Model & Dev & Number (Dev) & Test \\
        \hline
        TASE & 83.58 & 81.38 & 83.62 \\
        w/ CANTOR & \textbf{83.93} & \textbf{81.95} & \textbf{84.25}\\
        \hline
    \end{tabular}
    }
    \label{tab:drop}
    }
    \caption{F1 scores on DROP\textsubscript{num} and DROP. \textit{w/ CANTOR} is a TASE model that replaces the original tagging-based arithmetic module with CANTOR; all modules share one problem encoder.}
    \label{tab:drop_full}
\end{table}
\endgroup

\noindent
CANTOR is also applicable to weakly-supervised scenarios where only final answers are annotated.
Given problem-answer pairs $\{\langle X, A \rangle \}$, if it is feasible to find $Y$ that evaluates to $A$, we can adapt hard EM, MML, and hard EM with annealing for weakly-supervised training by simply re-defining $\Gamma$ for the objective functions as follows:
\begin{equation}
    \small
    \begin{split}
        \Gamma = \{Z|\exists Y\ P(A|Y)P_\theta(Y|Z, X)=1 \}
    \end{split}\nonumber
\end{equation}
where $P(A|Y)$ is 1 if and only if $Y$ evaluates to $A$.

For weakly-supervised training on DROP\textsubscript{num}, we followed TASE to enumerate $Y$ by searching addition or subtraction of two numbers, and used MML for training\footnote{There are more advanced weakly-supervised training methods \cite{DBLP:conf/iclr/ChenLYZSL20,DBLP:conf/acl/ShaoSLH20} for discrete reasoning on DROP. Investigation of how CANTOR is compatible with them is left for future work.}. As each $Y$ has only one operation, MML conducts exact marginalization over $\Gamma$.

As shown by Table \ref{tab:drop_num}, CANTOR significantly outperforms TASE\textsubscript{arith} on DROP\textsubscript{num}.
If using CANTOR as a drop-in replacement for the arithmetic module of TASE, we can obtain further improvements on DROP (Table \ref{tab:drop}).

\subsection{Ablation Study}
\subsubsection{No Pre-defined Order Restrictions}
To investigate the effect of removing restrictions on decoding dependencies, we considered a variant of CANTOR called \textbf{vanilla CANTOR}, which also produces all operations in parallel, but is not designed to have diverse and possibly redundant operations for comparisons in both operand matching and root vertex selection.
Specifically, instead of using a pre-specified value of $L$, vanilla CANTOR predicts the number of operations needed to solve a given problem as $L$ (using the \texttt{[CLS]} representation from the encoder), and was trained with na\"ive mapping; the last vertex $v_L$ is the root vertex.
As shown by Table \ref{tab:ablate_order_structure}, vanilla CANTOR already outperforms the best baseline which adopts a pre-defined decoding order, indicating that our model does well in capturing the structures of equations internally, and that using a pre-defined decoding order may be an unnecessary burden on model learning.

\begingroup
\setlength{\tabcolsep}{3pt} 
\renewcommand{\arraystretch}{1} 
\begin{table}[t!]
    \centering
	\adjustbox{max width=.45\textwidth}{
    \begin{tabular}{lcccccc}
        \hline
        \multirow{2}{*}{Model} & \multicolumn{2}{c}{MathQA (Dev)} & \multicolumn{2}{c}{MathQA (Test)} & \multicolumn{2}{c}{SVAMP} \\
        \cmidrule(lr){2-3} \cmidrule(lr){4-5} \cmidrule(lr){6-7}
        & Equ. & Val. & Equ. & Val. & Equ. & Val. \\
        \hline
        \multicolumn{7}{c}{\textit{Autoregressive Model}}\\
        \hline
        \multicolumn{7}{l}{\textit{Pre-defined Decoding Order (\checkmark); Structure Modeling (\checkmark)}} \\
        D\textsc{educt}R\textsc{easoner} & 74.0 & 77.5 & 74.7 & 78.6 & 44.6 & 45.1 \\
        \hline
        \multicolumn{7}{c}{\textit{Non-autoregressive Models}}\\
        \hline
        \multicolumn{7}{l}{\textit{Pre-defined Decoding Order ({\small \XSolidBrush}); Structure Modeling ({\small \XSolidBrush})}} \\
        Vanilla NAR & 76.9 & 79.1 & 77.4 & 79.6 & 36.4\textsubscript{$\pm 1.56$} & 37.0\textsubscript{$\pm 1.48$} \\
        \multicolumn{7}{l}{\textit{Pre-defined Decoding Order ({\small \XSolidBrush}); Structure Modeling (\checkmark)}} \\
        Vanilla CANTOR & \textbf{77.4} & \textbf{80.4} & \textbf{78.3} & \textbf{81.4} & \textbf{46.8\textsubscript{$\pm 0.55$}} & \textbf{47.3\textsubscript{$\pm 0.47$}} \\
        \hline
    \end{tabular}
    }
    \caption{Comparisons between (1) models without and with a pre-defined decoding order (Vanilla CANTOR vs. D\textsc{educt}R\textsc{easoner}) (2) and models with and without modeling the structures of equations (Vanilla CANTOR vs. Vanilla NAR).}
    \label{tab:ablate_order_structure}
\end{table}
\endgroup
\subsubsection{Structure Modeling}
Previous work has proposed non-autoregressive models for semantic parsing, but without explicit structure modeling.
To investigate the effect of structure modeling, we compared vanilla CANTOR with the non-autoregressive parser proposed by \citet{DBLP:conf/emnlp/ShrivastavaCBDA21} which we name as \textbf{vanilla NAR}.
Vanilla NAR predicts a length of the decoder $L^\prime$, and produces an $L^\prime$-sized equation text with token-wise generation (one token at a position)\footnote{For numbers in an equation, following \citet{DBLP:conf/emnlp/ShrivastavaCBDA21}, vanilla NAR decodes their positions in the problem instead of their constituent tokens. An example of an equation text is \texttt{( const\_1 + pos@7 ) $\times$ const\_2} where \texttt{pos@7} denotes the number mentioned at position 7.}.
By contrast, vanilla CANTOR structures an equation as a DAG with vertices corresponding to reasoning steps.
As shown by Table \ref{tab:ablate_order_structure}, vanilla CANTOR outperforms vanilla NAR, which verifies the value of structure modeling.

\begin{figure*}[t!]
    \centering
    \includegraphics[width=0.97\textwidth]{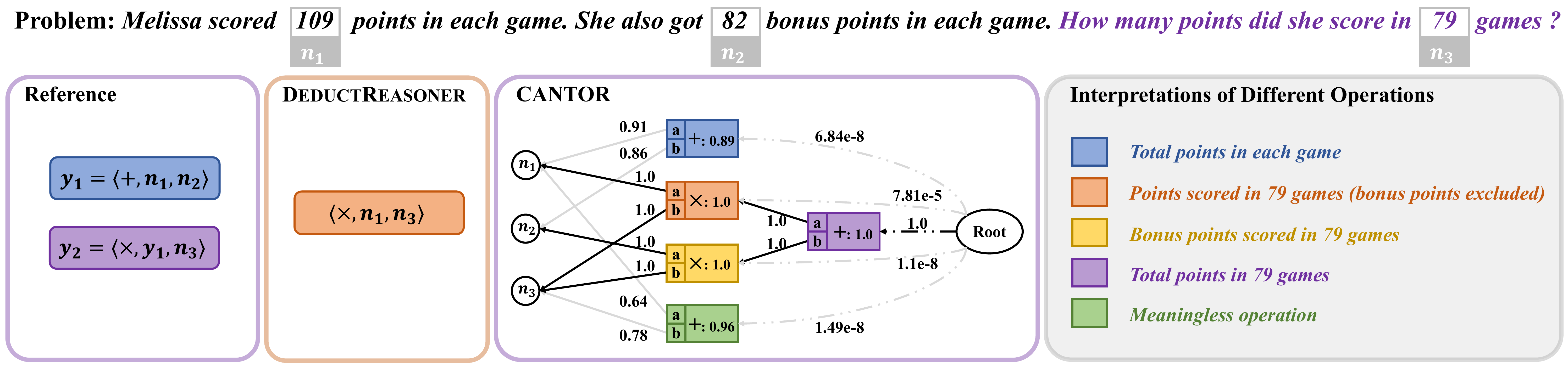}
    \caption{A test case from SVAMP. Operations leading to the same quantity are marked with the same color. \textcolor{purple}{Purple} ones are operations evaluating to the correct answer. For a clear presentation of our DAG, we only retain top-5 root vertices along with their descendants. We also present probabilities of predicted operators, operands, and root vertices.
    The best baseline D\textsc{educt}R\textsc{easoner} overlooks \textit{bonus points} in its prediction; while the same prediction appears as a sub-graph in our DAG, CANTOR succeeds in filtering it out and recognizes the correct one.}
    \label{fig:svamp_case}
\end{figure*}
\subsubsection{Capturing Diverse Reasoning Steps}
CANTOR decodes an $L$-sized DAG that encompasses diverse reasoning steps which are necessary or possibly redundant.
Comparing diverse choices is beneficial to pick out the proper one.
In this section, we investigate how well CANTOR captures diverse reasoning steps and its effect on model performance.
As it is pointless to merely have different operations at different vertices, we focused on the quality of top-$k$ root vertices ($\mbox{top-}k\ P_r(p_{|Z|}|X)$)\footnote{When selecting the top-$k$ root vertices, we skipped repeated logically-equivalent equations; logical equivalence is evaluated the same way as equation accuracy.\label{footnote:top-k}} and evaluated the recall of answers (\textit{Val.@$k$}).

\begingroup
\setlength{\tabcolsep}{3pt} 
\renewcommand{\arraystretch}{1} 
\begin{table}[t!]
    \centering
	\adjustbox{max width=.49\textwidth}{
    \begin{tabular}{ccccccc}
        \hline
        \multirow{2}{*}{Training Method} & \multicolumn{2}{c}{MathQA (Dev)} & \multicolumn{2}{c}{MathQA (Test)} & \multicolumn{2}{c}{SVAMP} \\
        \cmidrule(lr){2-3} \cmidrule(lr){4-5} \cmidrule(lr){6-7}
        & Val.@1 & Val.@5 & Val.@1 & Val.@5 & Val.@1 & Val.@5 \\
        \hline
        Na\"ive Mapping & 80.51 & 81.63 & 81.00 & 81.87 & 48.22\textsubscript{$\pm 0.94$} & 55.98\textsubscript{$\pm 1.30$} \\
        \hline
        Hard EM & 81.29 & 82.46 & 82.06 & 83.68 & 47.08\textsubscript{$\pm 0.63$} & 66.56\textsubscript{$\pm 1.45$}\\
        \hline
        MML & 68.39 & 71.09 & 69.91 & 72.65 & \textbf{49.58\textsubscript{$\pm 0.63$}} & 63.44\textsubscript{$\pm 1.38$} \\
        \hline
        \multicolumn{7}{l}{Hard EM with Annealing} \\
        $\tau = 500$ & 81.46 & \textbf{83.66} & \textbf{82.93} & \textbf{84.86} & 47.84\textsubscript{$\pm 0.64$} & \textbf{67.52\textsubscript{$\pm 1.78$}} \\
        $\tau = 1,000$ & 81.54 & 83.20 & 82.55 & 83.99 & 48.06\textsubscript{$\pm 0.91$} & 66.46\textsubscript{$\pm 2.95$} \\
        $\tau = 1,500$ & 81.50 & 83.37 & 82.68 & 84.24 & 48.82\textsubscript{$\pm 0.58$} & 67.12\textsubscript{$\pm 1.63$} \\
        $\tau = 2,000$ & \textbf{81.54} & 83.45 & 82.80 & 84.30 & 48.14\textsubscript{$\pm 0.48$} & 65.58\textsubscript{$\pm 1.79$}\\
        \hline
    \end{tabular}
    }
    \caption{Comparisons among different training methods.
    \textit{Val.@$k$} is the recall of answers over execution results at top-$k$ root vertices ($\mbox{top-}k\ P_r(p_{|Z|}|X)$).}
    \label{tab:ablate_training_methods}
\end{table}
\endgroup
\noindent
\textbf{Training Methods}\label{sec:ablate_training_methods}
Compared with vanilla CANTOR, CANTOR trained with methods that leverage more vertices than necessary (for ground-truth equations) achieved higher \textit{Val.@$k$} most of the time (Table \ref{tab:ablate_training_methods}).
One exception was applying MML on MathQA, which led to much worse performance.
We conjecture that this is because our assumption in MML is incompatible with the complex equations in MathQA (please refer to Appendix \ref{appendix:limit_mml} for detailed discussion on the limitations of our MML).
However, it is still helpful to warm up hard EM with MML, which is demonstrated by the improvements of hard EM with annealing over hard EM.
Notably, CANTOR trained with na\"ive mapping outperforms vanilla CANTOR on SVAMP; this is because the former was trained to leverage more vertices than necessary in testing (due to $\max |Y|$ on the train set being larger than $\max |Y|$ on SVAMP) and compares different vertices for root vertex selection, while the latter has no access to extra vertices and uses the last vertex as the root vertex without comparisons.
\begingroup
\setlength{\tabcolsep}{3pt} 
\renewcommand{\arraystretch}{1} 
\begin{table}[t!]
    \centering
	\adjustbox{max width=.43\textwidth}{
    \begin{tabular}{ccccccc}
        \hline
        \multirow{2}{*}{$L$} & \multicolumn{2}{c}{MathQA (Dev)} & \multicolumn{2}{c}{MathQA (Test)} & \multicolumn{2}{c}{SVAMP} \\
        \cmidrule(lr){2-3} \cmidrule(lr){4-5} \cmidrule(lr){6-7}
        & Val.@1 & Val.@5 & Val.@1 & Val.@5 & Val.@1 & Val.@5 \\
        \hline
        20 & \cellcolor{orange!5}81.00 & \cellcolor{orange!5}82.66 & \cellcolor{orange!5}82.74 & \cellcolor{orange!5}83.86 & \cellcolor{orange!5}48.56\textsubscript{$\pm 0.43$} & \cellcolor{orange!5}59.00\textsubscript{$\pm 2.28$}\\
        40 & \cellcolor{orange!15}81.21 & \cellcolor{orange!15}82.70 & \cellcolor{orange!5}82.74 & \cellcolor{orange!5}83.86 & \cellcolor{orange!30}48.96\textsubscript{$\pm 0.74$} & \cellcolor{orange!25}62.94\textsubscript{$\pm 2.26$} \\
        60 & \cellcolor{orange!25}81.54 & \cellcolor{orange!45}\textbf{83.45} & \cellcolor{orange!25}82.80 & \cellcolor{orange!25}84.30 & \cellcolor{orange!45}\textbf{49.58\textsubscript{$\pm 0.63$}} & \cellcolor{orange!45}\textbf{63.44\textsubscript{$\pm 1.38$}} \\
        80 & \cellcolor{orange!45}\textbf{81.67} & \cellcolor{orange!30}83.16 & \cellcolor{orange!45}\textbf{82.93} & \cellcolor{orange!45}\textbf{84.42} & \cellcolor{orange!20}48.58\textsubscript{$\pm 0.48$} & \cellcolor{orange!15}62.32\textsubscript{$\pm 0.72$} \\
        100 & \cellcolor{orange!35}81.58 & \cellcolor{orange!35}83.20 & \cellcolor{orange!35}82.87 & \cellcolor{orange!35}84.42 & \cellcolor{orange!25}48.60\textsubscript{$\pm 0.75$} & \cellcolor{orange!35}63.24\textsubscript{$\pm 1.87$} \\
        \hline
    \end{tabular}
    }
    \caption{\textit{Val.@$k$} with varying graph sizes $L$. Models were trained using hard EM with annealing ($\tau=2000$) on MathQA and MML on SVAMP. \textit{Val.@$k$} is the answer recall over execution results at top-$k$ root vertices.}
    \label{tab:ablate_graph_size}
\end{table}
\endgroup

\textbf{\textit{In practice, hard EM with annealing should be the default training method;}} as in Table \ref{tab:ablate_training_methods}, it always outperforms na\"ive mapping and hard EM, and is at least competitive with MML.
As shown by Table \ref{tab:ablate_training_methods} and Table \ref{tab:ablate_beam_size}, the two hyperparameters to tune, i.e., the number of warm-up steps $\tau$ and the beam size $B$, are robust to a wide range of values.

\noindent
\textbf{Graph Size $L$}\label{sec:ablate_graph_size}
A larger DAG can encompass more reasoning steps, but also increases the difficulty of operand matching and root vertex selection.
Training methods like hard EM may even suffer from suppressing false negative operations.
Table \ref{tab:ablate_graph_size} shows the effect of varying graph sizes $L$.
Model performance improves until $L$ reaches 80 and 60 on MathQA and SVAMP, respectively.

\subsection{Case Study}
Fig \ref{fig:svamp_case} presents a test case from SVAMP.
For a clear presentation of our DAG, we only show top-5 root vertices\textsuperscript{\ref{footnote:top-k}} along with their descendants.
By comparing diverse operations and chaining relevant ones, CANTOR succeeds in discriminating logically correct operations from distracting ones (e.g., the one predicted by D\textsc{educt}R\textsc{easoner} which overlooks \textit{bonus points}), even though the final equation is structurally different from the annotated reference.

\subsection{CANTOR vs. LLMs with Chain-of-Thought Prompting}
\begingroup
\setlength{\tabcolsep}{3pt} 
\renewcommand{\arraystretch}{1} 
\begin{table}[t!]
    \centering
	\adjustbox{max width=.45\textwidth}{
    \begin{tabular}{lccc}
        \hline
        Model & Params & SVAMP & GSM8K \\
        \hline
        \multicolumn{4}{c}{Few-Shot Setting}\\
        \hline
        \multicolumn{4}{l}{8-shot CoT \cite{DBLP:journals/corr/abs-2201-11903}} \\
        \multicolumn{1}{c}{\textit{LaMDA}} & 137B & 37.5 & 14.3 \\
        \multicolumn{1}{c}{\textit{GPT-3}} & 175B & 68.9 & 46.9 \\
        \multicolumn{1}{c}{\textit{PaLM}} & 62B & 46.7 & 29.9 \\
        & 540B & 79.0 & 56.9 \\
        \hline
        \multicolumn{4}{c}{Fully-Supervised Setting}\\
        \hline
        GPT-3 \cite{DBLP:journals/corr/abs-2110-14168} & 175B & - & $\sim$35 \\
        \hline
        \multicolumn{4}{l}{D\textsc{educt}R\textsc{easoner}} \\
        \multicolumn{1}{c}{\textit{RoBERTa\textsubscript{base}}} & 125M & 45.1 & - \\
        \multicolumn{1}{c}{\textit{RoBERTa\textsubscript{large}}} & 355M & 50.4 & - \\
        \hline
        \multicolumn{4}{l}{CANTOR} \\
        \multicolumn{1}{c}{\textit{RoBERTa\textsubscript{base}}} & 125M & 49.6 & - \\
        \multicolumn{1}{c}{\textit{RoBERTa\textsubscript{large}}} & 355M & 55.4 & 30.2 \\
        \hline
    \end{tabular}
    }
    \caption{Value accuracy on SVAMP and GSM8K. CoT is short for Chain-of-Thought prompting.}
    \label{tab:vs_cot}
\end{table}
\endgroup

Recently, \citet{DBLP:journals/corr/abs-2201-11903} proposed chain-of-thought prompting which endows large language models with the ability to generate a series of intermediate reasoning steps to reach the final answer of a given problem, achieving state-of-the-art performance on a wide range of reasoning tasks.
Table \ref{tab:vs_cot} compares CANTOR and chain-of-thought prompting.
Though being 392$\times$ smaller, CANTOR with RoBERTa\textsubscript{base} already outperforms PaLM-62B on SVAMP; using RoBERTa\textsubscript{large} gives an aggressive improvement, demonstrating CANTOR's great potential.

CANTOR is also applicable to the challenging GSM8K dataset \cite{DBLP:journals/corr/abs-2110-14168} which was created to probe the reasoning ability of large language models and has high diversity among problems.
As GSM8K was annotated with natural language solutions, the extracted equations are noisy and incomplete; we ended up with 6,312 (out of 7,473) noisy training examples.
As shown in Table \ref{tab:vs_cot}, CANTOR is close to the 175B GPT-3 model fine-tuned on the whole train set, and is on a par with PaLM-62B with chain-of-thought prompting.



\section{Conclusion}
We propose a numerical reasoner called CANTOR.
Unlike previous structured decoders that model a single equation with pre-defined restrictions on the decoding dependencies, CANTOR models diverse reasoning steps using a directed acyclic graph without a pre-defined decoding order, and derives equations by comparing and chaining relevant reasoning steps.
With our training methods, CANTOR is capable of capturing the logical dependencies among reasoning steps internally, and produces equations that are more consistent with the reasoning problems by comparing diverse reasoning steps.
CANTOR achieves state-of-the-art results on two math word problem datasets under the fully-supervised setting, and is applicable to weakly-supervised scenarios with significant improvements.

In future work, we plan to extend CANTOR for general structured prediction tasks, e.g., sequence labeling and parsing.

\section{Limitations}
Though CANTOR significantly outperforms baselines, there is still a large room for improvement in solving numerical reasoning problems with novel equation templates and being robust to variations in the problems.
For example, our value accuracy on SVAMP problems with unseen equation templates is lower than 20\% (Table \ref{tab:mwp_breakdown}), and the value accuracy on problems that evaluate question sensitivity barely reaches 30\% (Table \ref{tab:svamp_breakdown_variations}).
We also argue for more benchmarks that expose weaknesses of existing models, as we observe that more than half of test problems in MWP datasets can be solved with equation templates seen in training, which may overestimate the numerical reasoning ability of neural models.

\section*{Acknowledgements}
This work was supported by the National Science Foundation for Distinguished Young Scholars (with No. 62125604) and the NSFC projects (Key project with No. 61936010 and regular project with No. 61876096). This work was also supported by the Guoqiang Institute of Tsinghua University, with Grant No. 2019GQG1 and 2020GQG0005, and sponsored by Tsinghua-Toyota Joint Research Fund.

\bibliography{dataset,NAR_parser,NAR_translation,reranking,sequential_decoding,structured_decoding,DROP_solver,PLMs,custom}

\begin{thebibliography}{38}
\expandafter\ifx\csname natexlab\endcsname\relax\def\natexlab#1{#1}\fi

\bibitem[{Amini et~al.(2019)Amini, Gabriel, Lin, Koncel{-}Kedziorski, Choi, and
  Hajishirzi}]{DBLP:conf/naacl/AminiGLKCH19}
Aida Amini, Saadia Gabriel, Shanchuan Lin, Rik Koncel{-}Kedziorski, Yejin Choi,
  and Hannaneh Hajishirzi. 2019.
\newblock \href {https://doi.org/10.18653/v1/n19-1245} {Mathqa: Towards
  interpretable math word problem solving with operation-based formalisms}.
\newblock In \emph{Proceedings of the 2019 Conference of the North American
  Chapter of the Association for Computational Linguistics: Human Language
  Technologies, {NAACL-HLT} 2019, Minneapolis, MN, USA, June 2-7, 2019, Volume
  1 (Long and Short Papers)}, pages 2357--2367. Association for Computational
  Linguistics.

\bibitem[{Babu et~al.(2021)Babu, Shrivastava, Aghajanyan, Aly, Fan, and
  Ghazvininejad}]{DBLP:conf/naacl/BabuSAAFG21}
Arun Babu, Akshat Shrivastava, Armen Aghajanyan, Ahmed Aly, Angela Fan, and
  Marjan Ghazvininejad. 2021.
\newblock \href {https://doi.org/10.18653/v1/2021.naacl-main.236}
  {Non-autoregressive semantic parsing for compositional task-oriented dialog}.
\newblock In \emph{Proceedings of the 2021 Conference of the North American
  Chapter of the Association for Computational Linguistics: Human Language
  Technologies, {NAACL-HLT} 2021, Online, June 6-11, 2021}, pages 2969--2978.
  Association for Computational Linguistics.

\bibitem[{Cao et~al.(2021)Cao, Hong, Li, and Luo}]{DBLP:conf/aaai/CaoH0021}
Yixuan Cao, Feng Hong, Hongwei Li, and Ping Luo. 2021.
\newblock \href {https://ojs.aaai.org/index.php/AAAI/article/view/16075} {A
  bottom-up {DAG} structure extraction model for math word problems}.
\newblock In \emph{Thirty-Fifth {AAAI} Conference on Artificial Intelligence,
  {AAAI} 2021, Thirty-Third Conference on Innovative Applications of Artificial
  Intelligence, {IAAI} 2021, The Eleventh Symposium on Educational Advances in
  Artificial Intelligence, {EAAI} 2021, Virtual Event, February 2-9, 2021},
  pages 39--46. {AAAI} Press.

\bibitem[{Chen et~al.(2020)Chen, Liang, Yu, Zhou, Song, and
  Le}]{DBLP:conf/iclr/ChenLYZSL20}
Xinyun Chen, Chen Liang, Adams~Wei Yu, Denny Zhou, Dawn Song, and Quoc~V. Le.
  2020.
\newblock \href {https://openreview.net/forum?id=ryxjnREFwH} {Neural symbolic
  reader: Scalable integration of distributed and symbolic representations for
  reading comprehension}.
\newblock In \emph{8th International Conference on Learning Representations,
  {ICLR} 2020, Addis Ababa, Ethiopia, April 26-30, 2020}. OpenReview.net.

\bibitem[{Chen et~al.(2021)Chen, Chen, Smiley, Shah, Borova, Langdon, Moussa,
  Beane, Huang, Routledge, and Wang}]{DBLP:conf/emnlp/ChenCSSBLMBHRW21}
Zhiyu Chen, Wenhu Chen, Charese Smiley, Sameena Shah, Iana Borova, Dylan
  Langdon, Reema Moussa, Matt Beane, Ting{-}Hao Huang, Bryan~R. Routledge, and
  William~Yang Wang. 2021.
\newblock \href {https://doi.org/10.18653/v1/2021.emnlp-main.300} {Finqa: {A}
  dataset of numerical reasoning over financial data}.
\newblock In \emph{Proceedings of the 2021 Conference on Empirical Methods in
  Natural Language Processing, {EMNLP} 2021, Virtual Event / Punta Cana,
  Dominican Republic, 7-11 November, 2021}, pages 3697--3711. Association for
  Computational Linguistics.

\bibitem[{Chowdhery et~al.(2022)Chowdhery, Narang, Devlin, Bosma, Mishra,
  Roberts, Barham, Chung, Sutton, Gehrmann, Schuh, Shi, Tsvyashchenko, Maynez,
  Rao, Barnes, Tay, Shazeer, Prabhakaran, Reif, Du, Hutchinson, Pope, Bradbury,
  Austin, Isard, Gur{-}Ari, Yin, Duke, Levskaya, Ghemawat, Dev, Michalewski,
  Garcia, Misra, Robinson, Fedus, Zhou, Ippolito, Luan, Lim, Zoph, Spiridonov,
  Sepassi, Dohan, Agrawal, Omernick, Dai, Pillai, Pellat, Lewkowycz, Moreira,
  Child, Polozov, Lee, Zhou, Wang, Saeta, Diaz, Firat, Catasta, Wei,
  Meier{-}Hellstern, Eck, Dean, Petrov, and
  Fiedel}]{DBLP:journals/corr/abs-2204-02311}
Aakanksha Chowdhery, Sharan Narang, Jacob Devlin, Maarten Bosma, Gaurav Mishra,
  Adam Roberts, Paul Barham, Hyung~Won Chung, Charles Sutton, Sebastian
  Gehrmann, Parker Schuh, Kensen Shi, Sasha Tsvyashchenko, Joshua Maynez,
  Abhishek Rao, Parker Barnes, Yi~Tay, Noam Shazeer, Vinodkumar Prabhakaran,
  Emily Reif, Nan Du, Ben Hutchinson, Reiner Pope, James Bradbury, Jacob
  Austin, Michael Isard, Guy Gur{-}Ari, Pengcheng Yin, Toju Duke, Anselm
  Levskaya, Sanjay Ghemawat, Sunipa Dev, Henryk Michalewski, Xavier Garcia,
  Vedant Misra, Kevin Robinson, Liam Fedus, Denny Zhou, Daphne Ippolito, David
  Luan, Hyeontaek Lim, Barret Zoph, Alexander Spiridonov, Ryan Sepassi, David
  Dohan, Shivani Agrawal, Mark Omernick, Andrew~M. Dai,
  Thanumalayan~Sankaranarayana Pillai, Marie Pellat, Aitor Lewkowycz, Erica
  Moreira, Rewon Child, Oleksandr Polozov, Katherine Lee, Zongwei Zhou, Xuezhi
  Wang, Brennan Saeta, Mark Diaz, Orhan Firat, Michele Catasta, Jason Wei,
  Kathy Meier{-}Hellstern, Douglas Eck, Jeff Dean, Slav Petrov, and Noah
  Fiedel. 2022.
\newblock \href {https://doi.org/10.48550/arXiv.2204.02311} {Palm: Scaling
  language modeling with pathways}.
\newblock \emph{CoRR}, abs/2204.02311.

\bibitem[{Cobbe et~al.(2021)Cobbe, Kosaraju, Bavarian, Hilton, Nakano, Hesse,
  and Schulman}]{DBLP:journals/corr/abs-2110-14168}
Karl Cobbe, Vineet Kosaraju, Mohammad Bavarian, Jacob Hilton, Reiichiro Nakano,
  Christopher Hesse, and John Schulman. 2021.
\newblock \href {http://arxiv.org/abs/2110.14168} {Training verifiers to solve
  math word problems}.
\newblock \emph{CoRR}, abs/2110.14168.

\bibitem[{Devlin et~al.(2019)Devlin, Chang, Lee, and
  Toutanova}]{DBLP:conf/naacl/DevlinCLT19}
Jacob Devlin, Ming{-}Wei Chang, Kenton Lee, and Kristina Toutanova. 2019.
\newblock \href {https://doi.org/10.18653/v1/n19-1423} {{BERT:} pre-training of
  deep bidirectional transformers for language understanding}.
\newblock In \emph{Proceedings of the 2019 Conference of the North American
  Chapter of the Association for Computational Linguistics: Human Language
  Technologies, {NAACL-HLT} 2019, Minneapolis, MN, USA, June 2-7, 2019, Volume
  1 (Long and Short Papers)}, pages 4171--4186. Association for Computational
  Linguistics.

\bibitem[{Du et~al.(2021)Du, Tu, and Jiang}]{DBLP:conf/icml/DuTJ21}
Cunxiao Du, Zhaopeng Tu, and Jing Jiang. 2021.
\newblock \href {http://proceedings.mlr.press/v139/du21c.html} {Order-agnostic
  cross entropy for non-autoregressive machine translation}.
\newblock In \emph{Proceedings of the 38th International Conference on Machine
  Learning, {ICML} 2021, 18-24 July 2021, Virtual Event}, volume 139 of
  \emph{Proceedings of Machine Learning Research}, pages 2849--2859. {PMLR}.

\bibitem[{Dua et~al.(2019)Dua, Wang, Dasigi, Stanovsky, Singh, and
  Gardner}]{DBLP:conf/naacl/DuaWDSS019}
Dheeru Dua, Yizhong Wang, Pradeep Dasigi, Gabriel Stanovsky, Sameer Singh, and
  Matt Gardner. 2019.
\newblock \href {https://doi.org/10.18653/v1/n19-1246} {{DROP:} {A} reading
  comprehension benchmark requiring discrete reasoning over paragraphs}.
\newblock In \emph{Proceedings of the 2019 Conference of the North American
  Chapter of the Association for Computational Linguistics: Human Language
  Technologies, {NAACL-HLT} 2019, Minneapolis, MN, USA, June 2-7, 2019, Volume
  1 (Long and Short Papers)}, pages 2368--2378. Association for Computational
  Linguistics.

\bibitem[{Ghazvininejad et~al.(2020)Ghazvininejad, Karpukhin, Zettlemoyer, and
  Levy}]{DBLP:conf/icml/GhazvininejadKZ20}
Marjan Ghazvininejad, Vladimir Karpukhin, Luke Zettlemoyer, and Omer Levy.
  2020.
\newblock \href {http://proceedings.mlr.press/v119/ghazvininejad20a.html}
  {Aligned cross entropy for non-autoregressive machine translation}.
\newblock In \emph{Proceedings of the 37th International Conference on Machine
  Learning, {ICML} 2020, 13-18 July 2020, Virtual Event}, volume 119 of
  \emph{Proceedings of Machine Learning Research}, pages 3515--3523. {PMLR}.

\bibitem[{Gu et~al.(2018)Gu, Bradbury, Xiong, Li, and
  Socher}]{DBLP:conf/iclr/Gu0XLS18}
Jiatao Gu, James Bradbury, Caiming Xiong, Victor O.~K. Li, and Richard Socher.
  2018.
\newblock \href {https://openreview.net/forum?id=B1l8BtlCb} {Non-autoregressive
  neural machine translation}.
\newblock In \emph{6th International Conference on Learning Representations,
  {ICLR} 2018, Vancouver, BC, Canada, April 30 - May 3, 2018, Conference Track
  Proceedings}. OpenReview.net.

\bibitem[{Hochreiter and Schmidhuber(1997)}]{DBLP:journals/neco/HochreiterS97}
Sepp Hochreiter and J{\"{u}}rgen Schmidhuber. 1997.
\newblock \href {https://doi.org/10.1162/neco.1997.9.8.1735} {Long short-term
  memory}.
\newblock \emph{Neural Comput.}, 9(8):1735--1780.

\bibitem[{Huang et~al.(2022)Huang, Zhou, Liu, Li, and
  Huang}]{DBLP:journals/corr/abs-2205-07459}
Fei Huang, Hao Zhou, Yang Liu, Hang Li, and Minlie Huang. 2022.
\newblock \href {https://doi.org/10.48550/arXiv.2205.07459} {Directed acyclic
  transformer for non-autoregressive machine translation}.
\newblock \emph{CoRR}, abs/2205.07459.

\bibitem[{Jie et~al.(2022)Jie, Li, and Lu}]{DBLP:conf/acl/JieLL22}
Zhanming Jie, Jierui Li, and Wei Lu. 2022.
\newblock \href {https://aclanthology.org/2022.acl-long.410} {Learning to
  reason deductively: Math word problem solving as complex relation
  extraction}.
\newblock In \emph{Proceedings of the 60th Annual Meeting of the Association
  for Computational Linguistics (Volume 1: Long Papers), {ACL} 2022, Dublin,
  Ireland, May 22-27, 2022}, pages 5944--5955. Association for Computational
  Linguistics.

\bibitem[{Koncel{-}Kedziorski et~al.(2016)Koncel{-}Kedziorski, Roy, Amini,
  Kushman, and Hajishirzi}]{DBLP:conf/naacl/Koncel-Kedziorski16}
Rik Koncel{-}Kedziorski, Subhro Roy, Aida Amini, Nate Kushman, and Hannaneh
  Hajishirzi. 2016.
\newblock \href {https://doi.org/10.18653/v1/n16-1136} {{MAWPS:} {A} math word
  problem repository}.
\newblock In \emph{{NAACL} {HLT} 2016, The 2016 Conference of the North
  American Chapter of the Association for Computational Linguistics: Human
  Language Technologies, San Diego California, USA, June 12-17, 2016}, pages
  1152--1157. The Association for Computational Linguistics.

\bibitem[{Lan et~al.(2021)Lan, Wang, Zhang, Lan, Dai, Wang, Zhang, and
  Lim}]{DBLP:journals/corr/abs-2109-00799}
Yihuai Lan, Lei Wang, Qiyuan Zhang, Yunshi Lan, Bing~Tian Dai, Yan Wang,
  Dongxiang Zhang, and Ee{-}Peng Lim. 2021.
\newblock \href {http://arxiv.org/abs/2109.00799} {Mwptoolkit: An open-source
  framework for deep learning-based math word problem solvers}.
\newblock \emph{CoRR}, abs/2109.00799.

\bibitem[{Li et~al.(2022)Li, Zhang, Yan, Zhou, Li, Liu, and
  Cao}]{DBLP:conf/acl/LiZYZLLC22}
Zhongli Li, Wenxuan Zhang, Chao Yan, Qingyu Zhou, Chao Li, Hongzhi Liu, and
  Yunbo Cao. 2022.
\newblock \href {https://aclanthology.org/2022.findings-acl.195} {Seeking
  patterns, not just memorizing procedures: Contrastive learning for solving
  math word problems}.
\newblock In \emph{Findings of the Association for Computational Linguistics:
  {ACL} 2022, Dublin, Ireland, May 22-27, 2022}, pages 2486--2496. Association
  for Computational Linguistics.

\bibitem[{Liang et~al.(2021)Liang, Zhang, Shao, and
  Zhang}]{DBLP:journals/corr/abs-2107-13435}
Zhenwen Liang, Jipeng Zhang, Jie Shao, and Xiangliang Zhang. 2021.
\newblock \href {http://arxiv.org/abs/2107.13435} {{MWP-BERT:} {A} strong
  baseline for math word problems}.
\newblock \emph{CoRR}, abs/2107.13435.

\bibitem[{Liu et~al.(2019)Liu, Ott, Goyal, Du, Joshi, Chen, Levy, Lewis,
  Zettlemoyer, and Stoyanov}]{DBLP:journals/corr/abs-1907-11692}
Yinhan Liu, Myle Ott, Naman Goyal, Jingfei Du, Mandar Joshi, Danqi Chen, Omer
  Levy, Mike Lewis, Luke Zettlemoyer, and Veselin Stoyanov. 2019.
\newblock \href {http://arxiv.org/abs/1907.11692} {Roberta: {A} robustly
  optimized {BERT} pretraining approach}.
\newblock \emph{CoRR}, abs/1907.11692.

\bibitem[{Miao et~al.(2020)Miao, Liang, and Su}]{DBLP:conf/acl/MiaoLS20}
Shen{-}Yun Miao, Chao{-}Chun Liang, and Keh{-}Yih Su. 2020.
\newblock \href {https://doi.org/10.18653/v1/2020.acl-main.92} {A diverse
  corpus for evaluating and developing english math word problem solvers}.
\newblock In \emph{Proceedings of the 58th Annual Meeting of the Association
  for Computational Linguistics, {ACL} 2020, Online, July 5-10, 2020}, pages
  975--984. Association for Computational Linguistics.

\bibitem[{Miller et~al.(1997)Miller, Stone, and Cox}]{599256}
M.L. Miller, H.S. Stone, and I.J. Cox. 1997.
\newblock \href {https://doi.org/10.1109/7.599256} {Optimizing murty's ranked
  assignment method}.
\newblock \emph{IEEE Transactions on Aerospace and Electronic Systems},
  33(3):851--862.

\bibitem[{Min et~al.(2019)Min, Chen, Hajishirzi, and
  Zettlemoyer}]{DBLP:conf/emnlp/MinCHZ19}
Sewon Min, Danqi Chen, Hannaneh Hajishirzi, and Luke Zettlemoyer. 2019.
\newblock \href {https://doi.org/10.18653/v1/D19-1284} {A discrete hard {EM}
  approach for weakly supervised question answering}.
\newblock In \emph{Proceedings of the 2019 Conference on Empirical Methods in
  Natural Language Processing and the 9th International Joint Conference on
  Natural Language Processing, {EMNLP-IJCNLP} 2019, Hong Kong, China, November
  3-7, 2019}, pages 2851--2864. Association for Computational Linguistics.

\bibitem[{Mishra et~al.(2022)Mishra, Mitra, Varshney, Sachdeva, Clark, Baral,
  and Kalyan}]{DBLP:conf/acl/MishraMVSCBK22}
Swaroop Mishra, Arindam Mitra, Neeraj Varshney, Bhavdeep~Singh Sachdeva, Peter
  Clark, Chitta Baral, and Ashwin Kalyan. 2022.
\newblock \href {https://aclanthology.org/2022.acl-long.246} {Numglue: {A}
  suite of fundamental yet challenging mathematical reasoning tasks}.
\newblock In \emph{Proceedings of the 60th Annual Meeting of the Association
  for Computational Linguistics (Volume 1: Long Papers), {ACL} 2022, Dublin,
  Ireland, May 22-27, 2022}, pages 3505--3523. Association for Computational
  Linguistics.

\bibitem[{Patel et~al.(2021)Patel, Bhattamishra, and
  Goyal}]{DBLP:conf/naacl/PatelBG21}
Arkil Patel, Satwik Bhattamishra, and Navin Goyal. 2021.
\newblock \href {https://doi.org/10.18653/v1/2021.naacl-main.168} {Are {NLP}
  models really able to solve simple math word problems?}
\newblock In \emph{Proceedings of the 2021 Conference of the North American
  Chapter of the Association for Computational Linguistics: Human Language
  Technologies, {NAACL-HLT} 2021, Online, June 6-11, 2021}, pages 2080--2094.
  Association for Computational Linguistics.

\bibitem[{Radford et~al.(2019)Radford, Wu, Child, Luan, Amodei, and
  Sutskever}]{Radford2019LanguageMA}
Alec Radford, Jeff Wu, Rewon Child, David Luan, Dario Amodei, and Ilya
  Sutskever. 2019.
\newblock Language models are unsupervised multitask learners.

\bibitem[{Ravichander et~al.(2019)Ravichander, Naik, Ros{\'{e}}, and
  Hovy}]{DBLP:conf/conll/RavichanderNRH19}
Abhilasha Ravichander, Aakanksha Naik, Carolyn~Penstein Ros{\'{e}}, and
  Eduard~H. Hovy. 2019.
\newblock \href {https://doi.org/10.18653/v1/K19-1033} {{EQUATE:} {A} benchmark
  evaluation framework for quantitative reasoning in natural language
  inference}.
\newblock In \emph{Proceedings of the 23rd Conference on Computational Natural
  Language Learning, CoNLL 2019, Hong Kong, China, November 3-4, 2019}, pages
  349--361. Association for Computational Linguistics.

\bibitem[{Segal et~al.(2020)Segal, Efrat, Shoham, Globerson, and
  Berant}]{DBLP:conf/emnlp/SegalESGB20}
Elad Segal, Avia Efrat, Mor Shoham, Amir Globerson, and Jonathan Berant. 2020.
\newblock \href {https://doi.org/10.18653/v1/2020.emnlp-main.248} {A simple and
  effective model for answering multi-span questions}.
\newblock In \emph{Proceedings of the 2020 Conference on Empirical Methods in
  Natural Language Processing, {EMNLP} 2020, Online, November 16-20, 2020},
  pages 3074--3080. Association for Computational Linguistics.

\bibitem[{Shao et~al.(2021)Shao, Shang, Liu, and
  Huang}]{DBLP:conf/acl/ShaoSLH20}
Zhihong Shao, Lifeng Shang, Qun Liu, and Minlie Huang. 2021.
\newblock \href {https://doi.org/10.18653/v1/2021.acl-long.318} {A mutual
  information maximization approach for the spurious solution problem in weakly
  supervised question answering}.
\newblock In \emph{Proceedings of the 59th Annual Meeting of the Association
  for Computational Linguistics and the 11th International Joint Conference on
  Natural Language Processing, {ACL/IJCNLP} 2021, (Volume 1: Long Papers),
  Virtual Event, August 1-6, 2021}, pages 4111--4124. Association for
  Computational Linguistics.

\bibitem[{Shen et~al.(2021)Shen, Yin, Li, Shang, Jiang, Zhang, and
  Liu}]{DBLP:conf/emnlp/ShenYLSJ0021}
Jianhao Shen, Yichun Yin, Lin Li, Lifeng Shang, Xin Jiang, Ming Zhang, and Qun
  Liu. 2021.
\newblock \href {https://doi.org/10.18653/v1/2021.findings-emnlp.195} {Generate
  {\&} rank: {A} multi-task framework for math word problems}.
\newblock In \emph{Findings of the Association for Computational Linguistics:
  {EMNLP} 2021, Virtual Event / Punta Cana, Dominican Republic, 16-20 November,
  2021}, pages 2269--2279. Association for Computational Linguistics.

\bibitem[{Shen and Jin(2020)}]{DBLP:conf/coling/ShenJ20}
Yibin Shen and Cheqing Jin. 2020.
\newblock \href {https://doi.org/10.18653/v1/2020.coling-main.262} {Solving
  math word problems with multi-encoders and multi-decoders}.
\newblock In \emph{Proceedings of the 28th International Conference on
  Computational Linguistics, {COLING} 2020, Barcelona, Spain (Online), December
  8-13, 2020}, pages 2924--2934. International Committee on Computational
  Linguistics.

\bibitem[{Shrivastava et~al.(2021)Shrivastava, Chuang, Babu, Desai, Arora,
  Zotov, and Aly}]{DBLP:conf/emnlp/ShrivastavaCBDA21}
Akshat Shrivastava, Pierce Chuang, Arun Babu, Shrey Desai, Abhinav Arora,
  Alexander Zotov, and Ahmed Aly. 2021.
\newblock \href {https://doi.org/10.18653/v1/2021.findings-emnlp.161} {Span
  pointer networks for non-autoregressive task-oriented semantic parsing}.
\newblock In \emph{Findings of the Association for Computational Linguistics:
  {EMNLP} 2021, Virtual Event / Punta Cana, Dominican Republic, 16-20 November,
  2021}, pages 1873--1886. Association for Computational Linguistics.

\bibitem[{Tan et~al.(2021)Tan, Wang, Jiang, and
  Jiang}]{DBLP:journals/corr/abs-2105-08928}
Minghuan Tan, Lei Wang, Lingxiao Jiang, and Jing Jiang. 2021.
\newblock \href {http://arxiv.org/abs/2105.08928} {Investigating math word
  problems using pretrained multilingual language models}.
\newblock \emph{CoRR}, abs/2105.08928.

\bibitem[{Vaswani et~al.(2017)Vaswani, Shazeer, Parmar, Uszkoreit, Jones,
  Gomez, Kaiser, and Polosukhin}]{DBLP:conf/nips/VaswaniSPUJGKP17}
Ashish Vaswani, Noam Shazeer, Niki Parmar, Jakob Uszkoreit, Llion Jones,
  Aidan~N. Gomez, Lukasz Kaiser, and Illia Polosukhin. 2017.
\newblock \href
  {https://proceedings.neurips.cc/paper/2017/hash/3f5ee243547dee91fbd053c1c4a845aa-Abstract.html}
  {Attention is all you need}.
\newblock In \emph{Advances in Neural Information Processing Systems 30: Annual
  Conference on Neural Information Processing Systems 2017, December 4-9, 2017,
  Long Beach, CA, {USA}}, pages 5998--6008.

\bibitem[{Wang et~al.(2017)Wang, Liu, and Shi}]{DBLP:conf/emnlp/WangLS17}
Yan Wang, Xiaojiang Liu, and Shuming Shi. 2017.
\newblock \href {https://doi.org/10.18653/v1/d17-1088} {Deep neural solver for
  math word problems}.
\newblock In \emph{Proceedings of the 2017 Conference on Empirical Methods in
  Natural Language Processing, {EMNLP} 2017, Copenhagen, Denmark, September
  9-11, 2017}, pages 845--854. Association for Computational Linguistics.

\bibitem[{Wei et~al.(2022)Wei, Wang, Schuurmans, Bosma, Chi, Le, and
  Zhou}]{DBLP:journals/corr/abs-2201-11903}
Jason Wei, Xuezhi Wang, Dale Schuurmans, Maarten Bosma, Ed~H. Chi, Quoc Le, and
  Denny Zhou. 2022.
\newblock \href {http://arxiv.org/abs/2201.11903} {Chain of thought prompting
  elicits reasoning in large language models}.
\newblock \emph{CoRR}, abs/2201.11903.

\bibitem[{Xie and Sun(2019)}]{DBLP:conf/ijcai/XieS19}
Zhipeng Xie and Shichao Sun. 2019.
\newblock \href {https://doi.org/10.24963/ijcai.2019/736} {A goal-driven
  tree-structured neural model for math word problems}.
\newblock In \emph{Proceedings of the Twenty-Eighth International Joint
  Conference on Artificial Intelligence, {IJCAI} 2019, Macao, China, August
  10-16, 2019}, pages 5299--5305. ijcai.org.

\bibitem[{Zhang et~al.(2020)Zhang, Wang, Lee, Bin, Wang, Shao, and
  Lim}]{DBLP:conf/acl/ZhangWLBWSL20}
Jipeng Zhang, Lei Wang, Roy~Ka{-}Wei Lee, Yi~Bin, Yan Wang, Jie Shao, and
  Ee{-}Peng Lim. 2020.
\newblock \href {https://doi.org/10.18653/v1/2020.acl-main.362} {Graph-to-tree
  learning for solving math word problems}.
\newblock In \emph{Proceedings of the 58th Annual Meeting of the Association
  for Computational Linguistics, {ACL} 2020, Online, July 5-10, 2020}, pages
  3928--3937. Association for Computational Linguistics.

\end{thebibliography}
\bibliographystyle{acl_natbib}

\newpage
\appendix

\section{Implementation Details}
\begingroup
\setlength{\tabcolsep}{3pt} 
\renewcommand{\arraystretch}{1} 
\begin{table}[htb]
    \centering
	\adjustbox{max width=.45\textwidth}{
    \begin{tabular}{ccc}
        \hline
        & MWP Solving & Discrete Reasoning \\
        \hline
        Batch Size & 32 & 12 \\
        Learning Rate & 2e-5 & 5e-6 \\
        Learning Rate Warum-up Steps & 500 & 0 \\
        \hline
    \end{tabular}
    }
    \caption{Hyperparameters for training CANTOR.}
    \label{tab:train_conf}
\end{table}
\endgroup

\noindent
We trained CANTOR for up to 100k training steps for the MWP task and up to 20 epochs for the discrete reasoning task, using hyperparameters specified in Table \ref{tab:train_conf}.
All experiments were conducted with V100 GPUs.

\section{Training Methods}
\label{appendix:training}

\subsection{Hard EM}
The training objective of hard EM is formulated as:
\begin{equation}
    \small
    \begin{split}
        \mathcal{L} = - \log P_\theta(Z^*|X),\ Z^*=arg\max_{Z \in \Gamma} P_\theta(Z|X)
    \end{split}\nonumber
\end{equation}
where $\Gamma = \{Z|P_\theta(Y|Z, X)=1\}$.
For any $Z \in \Gamma$, we have $|Z|=|Y|$, that is,
\begin{equation}
    \small
    \begin{split}
        Z =& \{z_1, z_2, ..., z_{|Y|}\},\ z_i = \langle p_i, z_i^f, z_i^a, z_i^b \rangle \\
        s.t.\ \ &1 \le p_1 < p_2 < ... < p_{|Y|} \le L \\
        & z_i^f = y_i^f
    \end{split}\nonumber
\end{equation}
where $z_i$ is the operation $y_i$ mapped to the vertex at position $p_i$.

As $\{p_1, ..., p_{|Y|}\}$ defines a valid mapping from $Y$ to $Z$, finding $Z^*$ is equivalent to finding the optimal mapping $\{p_1, ..., p_{|Y|}\}$, which we search for via beam search.
For convenience of illustration, we define the level of an operation in $Y$ as the length of the longest path from its corresponding vertex in the DAG counterpart of $Y$ to a leaf vertex (which is a constant or a number mentioned in the problem).
Let $D_l$ be the set of indices of operations with the same level $l$. For any $Z \in \Gamma$, $P_\theta(Z|X)$ can be factorized as follows:
\begin{equation}
    \small
    \begin{split}
        P_\theta(Z|X) = P_r(p_{|Y|}|X) \prod_l \prod_{i \in D_l} P_z(z_i|p_i, X)
    \end{split}\nonumber
\end{equation}
Therefore, we can use beam search to approximately find the optimal mapping level-by-level.
To guarantee valid mappings, we restrict that
\begin{equation}
    \small
    \begin{split}
        \forall i \in D_l,\ \max \{p_j|j \in D_{l - 1}\} < p_i \le L - \sum_{s > l} |D_s|
    \end{split}\nonumber
\end{equation}
To find the $B$-best mappings from $D_l$ according to $\prod_{i \in D_l} P_z(z_i|p_i, X)$, we utilize an open-source implementation\footnote{https://github.com/motrom/fastmurty} of Murty's algorithm \cite{599256}, whose worse case complexity is $O(B|D_l|^3)$.

\subsection{MML}
The training objective of MML is formulated as:
\begin{equation}
    \small
    \begin{split}
        \mathcal{L} = -\log \sum_{Z \in \Gamma} P_\theta(Z|X)
    \end{split}\nonumber
\end{equation}
We adopt a strong (but risky) assumption so that we can use dynamic programming to marginalize $P_\theta(Z|X)$ in polynomial time.
Specifically, for any operation $y_i$, we assume that the two sub-graphs rooted at $y_i^a$ and $y_i^b$ respectively (in the DAG counterpart of $Y$) are independently mapped to $\{v_1, v_2, ..., v_L\}$.
Let $\mathbf{M}_{i,j}$ be the marginal probability of the sub-graph rooted at $y_i$ mapped to $\{v_1, ..., v_j\}$, and $G(y_i)$ is the set of indices of constituent operations in the sub-graph, then $\mathbf{M}_{i,j}$ is computed as (we omit $X$ for a brief presentation):
\begin{equation}
    \small
    \begin{split}
        &\mathbf{M}_{i, j} = \sum_{\substack{\{p_k|k \in G(y_i)\}\\ p_i = j}} \prod_{k \in G(y_i)} P_z(z_k|p_k) \\
        &P_z(z_k|p_k) = P_f(y_k^f|p_k) P_a(z_k^a|p_k) P_b(z_k^b|p_k)
    \end{split}\nonumber
\end{equation}
Based on our assumption, if $y_i^a = y_u \in Y$ and $y_i^b = y_v \in Y$, we have:
\begin{equation}
    \small
    \begin{split}
        \mathbf{M}_{i, j} = P_f(y_i^f|j) \sum_{p_u=1}^{j-1} \mathbf{M}_{u, p_u} P_a(z_u|j) \sum_{p_v=1}^{j-1} \mathbf{M}_{v, p_v} P_b(z_v|j)
    \end{split}\nonumber
\end{equation}
otherwise, we have:
\begin{equation}
    \small
    \begin{split}
        y_i^a \in \mathcal{C} \cup& \mathcal{N}, y_i^b = y_v \in Y: \\
        &\mathbf{M}_{i, j} = P_f(y_i^f|j) P_a(y_i^a|j) \sum_{p_v=1}^{j-1} \mathbf{M}_{v, p_v} P_b(z_v|j) \\
        y_i^a = y_u& \in Y, y_i^b \in \mathcal{C}\cup\mathcal{N}:\\
        &\mathbf{M}_{i, j} = P_f(y_i^f|j) P_b(y_i^b|j) \sum_{p_u=1}^{j-1} \mathbf{M}_{u, p_u} P_a(z_u|j) \\
        y_i^a, y_i^b \in &\mathcal{C}\cup\mathcal{N}: \\
        &\mathbf{M}_{i, j} = P_f(y_i^f|j) P_a(y_i^a|j) P_b(y_i^b|j)
    \end{split}\nonumber
\end{equation}
Finally, the training objective can be computed as:
\begin{equation}
    \small
    \begin{split}
        \mathcal{L} = - \log \sum_{j=1}^L P_r(j) \mathbf{M}_{|Y|,j}
    \end{split}\nonumber
\end{equation}
which takes $O(|Y|)$ parallel operations.

\section{Limitations of Our MML}
\label{appendix:limit_mml}
For our MML method, we impose an independence assumption for efficient marginalization of $P_\theta(Z|X)$ over all $Z$ that denote valid mappings from operations in $Y$ to decoding positions, but at the cost of failing to compute exact marginalization and giving a noisy training objective when the target equation $Y$ is complex, like those in MathQA.


%
%

\paragraph{When does Our MML Conduct Exact Marginalization? And What are the Effects on Model Performance?}

Our MML conducts exact marginalization only if $Y$ has a linear structure, i.e., $Y$ has no branches; we define a branch in $Y$ to be an operation taking another two operations as operands.
If $Y$ have branches, our MML will include the probability of invalid $Z$ where different operations share one decoding position, which may mislead a model.
As validated by Table \ref{tab:ablate_branch}, (a) our MML works well on test problems whose gold equations have no branches (\# Branch=0: value accuracy=83.8\%), even when equations are long (\# Branch=0 and \# Operation>=4: value accuracy=79.5\%); (b) However, it becomes poor if equations have more branches (\# Branch>=2: value accuracy=35.0\%).

Empirically, our MML works well when most equations are linear, and short equations are likely linear in existing datasets (e.g., SVAMP and DROP).
When target equations are complex, hard EM should be more suitable, but we can still benefit from using our MML for warming up.

\begingroup
\setlength{\tabcolsep}{3pt} 
\renewcommand{\arraystretch}{1} 
\begin{table}[t!]
    \centering
	\adjustbox{max width=.48\textwidth}{
    \begin{tabular}{cccccccccc}
        \hline
        \multirow{2}{*}{\# Branch}& \multicolumn{3}{c}{All} & \multicolumn{3}{c}{\# Operation $\le$ 3} & \multicolumn{3}{c}{\# Operation $\ge$ 4} \\
        \cmidrule(lr){2-4} \cmidrule(lr){5-7} \cmidrule(lr){8-10}
        & MML & Na\"ive & Hard EM & MML & Na\"ive & Hard EM & MML & Na\"ive & Hard EM \\
        \hline
        0 & \textbf{83.8} & 82.5 & 82.9 & \textbf{84.7} & 83.4 & 83.4 & 79.5 & 78.0 & \textbf{80.3} \\
        1 & 69.2 & 83.0 & \textbf{85.2} & 82.3 & 82.3 & \textbf{87.1} & 65.6 & 83.2 & \textbf{84.7} \\
        $\ge$ 2 & 35.0 & 73.1 & \textbf{73.5} & - & - & - & 35.0 & 73.1 & \textbf{73.5} \\
        \hline
    \end{tabular}
    }
    \caption{Value accuracy breakdown on the test set of MathQA w.r.t. the number of branches (\textit{\# Branch}) and the number of operations (\textit{\# Operation}) in annotated gold equations. \textit{Na\"ive} stands for na\"ive mapping.}
    \label{tab:ablate_branch}
\end{table}
\endgroup

\section{Ablation Study}

\subsection{Effect of Beam Size $B$ on Hard EM}\label{sec:ablate_beam_size}
\begingroup
\setlength{\tabcolsep}{3pt} 
\renewcommand{\arraystretch}{1} 
\begin{table}[htb]
    \centering
	\adjustbox{max width=.35\textwidth}{
    \begin{tabular}{ccccc}
        \hline
        \multirow{2}{*}{Training Method} & \multicolumn{2}{c}{Dev} & \multicolumn{2}{c}{Test} \\
        \cmidrule(lr){2-3} \cmidrule(lr){4-5}
        & Equ. & Val. & Equ. & Val. \\
        \hline
        Random Mapping & 18.08 & 19.20 & 17.76 & 18.44 \\
        \hline
        \multicolumn{2}{l}{Hard EM} \\
        $B=1$ & 77.93 & 81.13 & \textbf{78.75} & \textbf{82.24} \\
        $B=10$ & 77.64 & 81.17 & 78.38 & 81.99 \\
        $B=20$ & \textbf{78.43} & \textbf{81.29} & 78.63 & 82.06 \\
        \hline
    \end{tabular}
    }
    \caption{
    Value accuracy of models trained with hard EM using different beam sizes. \textit{Random Mapping} is a baseline which uses random $Z \in \Gamma$ for training.
    }
    \label{tab:ablate_beam_size}
\end{table}
\endgroup
\noindent
As shown by Table \ref{tab:ablate_beam_size}, model performance is insensitive to beam size when using hard EM on MathQA.
To investigate whether the choices of $Z$ matter for optimization, we considered a baseline called \textbf{random mapping}, which optimizes a model on random $Z \in \Gamma$.
We observed that hard EM outperforms random mapping substantially, indicating that beam search finds effective $Z$ for training.

\section{Inference Efficiency}
Due to non-autoregressive decoding, CANTOR is significantly faster than previous autoregressive baselines in terms of inference efficiency. For example, on a single V100 32G GPU, CANTOR achieves a 7$\times$ speedup over D\textsc{educt}R\textsc{easoner} on the dev set of MathQA.

\begin{figure*}[t!]
    \centering
    \includegraphics[width=0.95\textwidth]{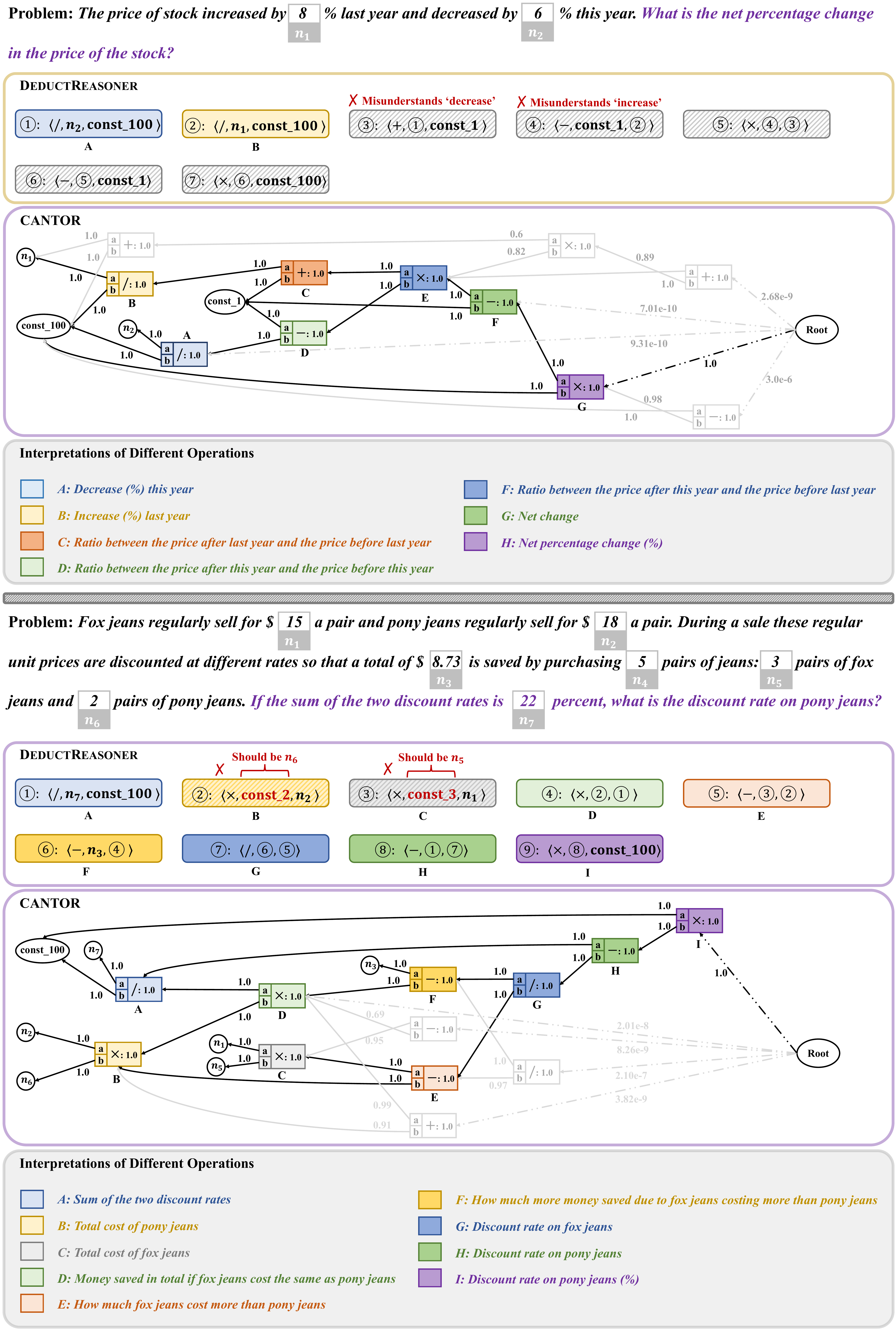}
    \caption{Two test cases from MathQA. Operations leading to the same value are marked with the same color and letter (e.g., A, B, etc.). \textcolor{purple}{Purple} ones are operations evaluating to the correct answer. For a clear presentation of our DAG, we only retain top-5 root vertices along with their descendants. We also present probabilities of predicted operators, operands, and root vertices.
    For predictions from D\textsc{educt}R\textsc{easoner}, we mark the decoding order of operations with circled numbers; operations with forward slashes in the background are erroneous ones.
    }
    \label{fig:mathqa_case}
\end{figure*}
\section{Case Study on MathQA}
\label{appendix:mathqa_case}
Fig \ref{fig:mathqa_case} presents two test cases from MathQA.
In the upper case, the baseline D\textsc{educt}R\textsc{easoner} misunderstands ``increase'' and ``decrease'', and conducts wrong operations.
In the lower case which mentions numerous quantities in the problem, D\textsc{educt}R\textsc{easoner}, despite arriving at the correct value, operates on wrong quantities at the second and the third reasoning steps.
By contrast, our proposed model CANTOR produces precise reasoning processes with proper choices of quantities to operate on.

\end{document}